%% file: riemann.tex
\documentclass[11pt,a4paper]{article}
\usepackage[utf8]{inputenc}
\usepackage[T1]{fontenc}
\usepackage{lmodern}
\usepackage{amsmath,amsfonts,amssymb,amsthm}
\usepackage{color}
\usepackage{graphicx}
\graphicspath{{./img/}}

\usepackage[showframe=false]{geometry}

\usepackage{url}
\usepackage{pdflscape}
\usepackage{afterpage}
\usepackage{capt-of}

\usepackage[svgnames,table]{xcolor}

\usepackage{longtable}
\usepackage{rotating}
\usepackage{array}
\usepackage{multicol}
\usepackage{multirow}

\usepackage[lined,boxed,commentsnumbered]{algorithm2e}

\usepackage[colorlinks=true,linkcolor=blue]{hyperref}

\usepackage{cleveref}

\input{header.tex}

\newcommand{\data}{\mathcal{D}}

\newcommand{\Norm}{\mathcal{N}}

\newcommand{\QD}{\mathrm{QD}}
\newcommand{\diag}{\mathrm{diag}}

\title{Practical Riemannian Neural Networks}
\author{Gaétan Marceau-Caron\and Yann Ollivier}
\date{\today}

\begin{document}

\maketitle

\begin{abstract}
We provide the
first experimental results on non-synthetic datasets for the
quasi-diagonal Riemannian
gradient descents
for neural networks introduced in \cite{Ollivier2015}. These include
the MNIST, SVHN, and FACE datasets as well as a previously
unpublished electroencephalogram dataset.
The
quasi-diagonal Riemannian algorithms consistently beat simple stochastic gradient
gradient descents by a varying margin. The computational overhead with
respect to simple backpropagation is around a factor $2$. Perhaps more
interestingly, these methods also
reach their final performance quickly, thus requiring fewer training
epochs and a smaller total computation time.

We also present an implementation guide to these Riemannian gradient descents
for neural networks, showing how the
quasi-diagonal versions can be implemented with minimal effort on top of
existing routines which compute gradients.
\end{abstract}

We present a practical and efficient implementation of invariant
stochastic gradient descent algorithms for neural networks based on
the \emph{quasi-diagonal} Riemannian metrics introduced in
\cite{Ollivier2015}. These can be implemented from the same data as
RMSProp- or AdaGrad-based schemes \cite{Duchi:11}, namely, by collecting gradients and
squared gradients for each data sample. Thus we will try to present them in a
way that can easily be incorporated on top of existing software providing
gradients for neural networks.

The main goal of these algorithms is to obtain invariance properties,
such as, for a neural network, insensitivity of the training algorithm to
whether a logistic or tanh activation function is used, or insensitivity
to simple changes of variables in the parameters, such as scaling some
parameters. Neither backpropagation nor AdaGrad-like schemes offer such
properties.

Invariance properties are important as they reduce the number of
arbitrary design choices, and guarantee that an observed good behavior in
one instance will transfer to other cases when, e.g., different scalings
are involved. In some cases this may alleviate the burden associated to
hyper-parameter tuning: in turn, sensible hyper-parameter values
sometimes become invariant.

Perhaps the most well-known invariant training procedure for statistical
learning is the \emph{natural gradient} promoted by Amari
\cite{Amari:1998}.  However, the natural gradient is rarely used in
practice for training neural networks.  The main reason is that the
Fisher information metric on which it relies is computationnally hard to
compute.  Different approximations have been proposed but can bring
little benefits compared to the implementation effort; see for instance
\cite{TONGA, Martens:14}. Moreover, it is not clear whether these
approximations preserve the invariance properties.\footnote{For instance,
the rank-one approximation in \cite{TONGA} is defined in reference to the Euclidean
norm of the difference between exact and approximated Fisher matrices,
and thus, is not parameterization-independent.}  In the end, simpler
techniques have been proposed to patch the flaws of the plain stochastic
gradient descent.  Using a balanced initialization with rectified linear
units, dropout and SGD or AdaGrad can be enough to obtain very good
results on many datasets. Nevertheless, these are tricks of the trade,
little justified from a mathematical point of view.



The Riemannian framework for neural networks 
\cite{Ollivier2015} allows us to define several \emph{quasi-diagonal}
metrics, which exactly keep some (but not all) invariance properties of
the natural gradient, at a smaller computational cost. The quasi-diagonal
structure means that the gradient obtained at each step of the
optimization is preconditioned by the inverse of a matrix which is almost
diagonal, with a few well-chosen non-diagonal terms that make it easy to
invert and have a special algebraic structure to ensure invariance with
respect to some simple transformations such as changing from sigmoid to
tanh activation function.

In this report, we assess the performance of these quasi-diagonal metrics
(only tested on synthetic data in \cite{Ollivier2015}). They turn out to
be quite competitive even after taking into account their computational
overhead (a factor of about $2$ with respect to simple backpropagation).
These methods consistently improve performance by various amounts depending on the
dataset, and also 
work well with dropout.
Especially, convergence is very fast early in the training procedure,
which opens the door to using fewer epochs and shorter overall training
times.

The code used in the experiments is available at\\
\url{https://www.lri.fr/~marceau/code/riemaNNv1.zip}

\paragraph{Invariance and gradient descents.}
The simplest instance of parameter invariance is that of parameter
scaling. Consider the classic gradient step
%
\begin{equation}
\label{eq:euclgrad}
\theta \leftarrow \theta - \eta \,\nabla_{\!\theta} f \qquad \text{(vanilla gradient descent)}
\end{equation}
where $\nabla_{\!\theta} f$ is the derivative of $f$ w.r.t.\ $\theta$.
This iteration scheme has an homogeneity problem in terms of physical
units: the unit of the derivative $\nabla_{\!\theta} f$ is the inverse of
the unit of $\theta$. This means that the learning rate $\eta$ itself
should have the homogeneity of $\theta^2$ for the gradient descent
\eqref{eq:euclgrad} to be
homogeneous. This is one reason why sensible learning rates can change a
lot from one problem to another.

AdaGrad-like methods \cite{Duchi:11} solve one half of this problem by rescaling
$\nabla_{\!\theta} f$ by its recent magnitude, so that it becomes of
order $1$ (physically dimensionless). The learning rate $\eta$ then has
the homogeneity of $\theta$, not $\theta^2$. This is particularly
relevant when it is known in advance that relevant parameter values for
$\theta$ are of order $1$ themselves; if not, the learning rate $\eta$
still needs to be set to a value compatible with the order of magnitude
of $\theta$.

\begin{figure}[t]
  \centering
  \includegraphics[scale=0.35]{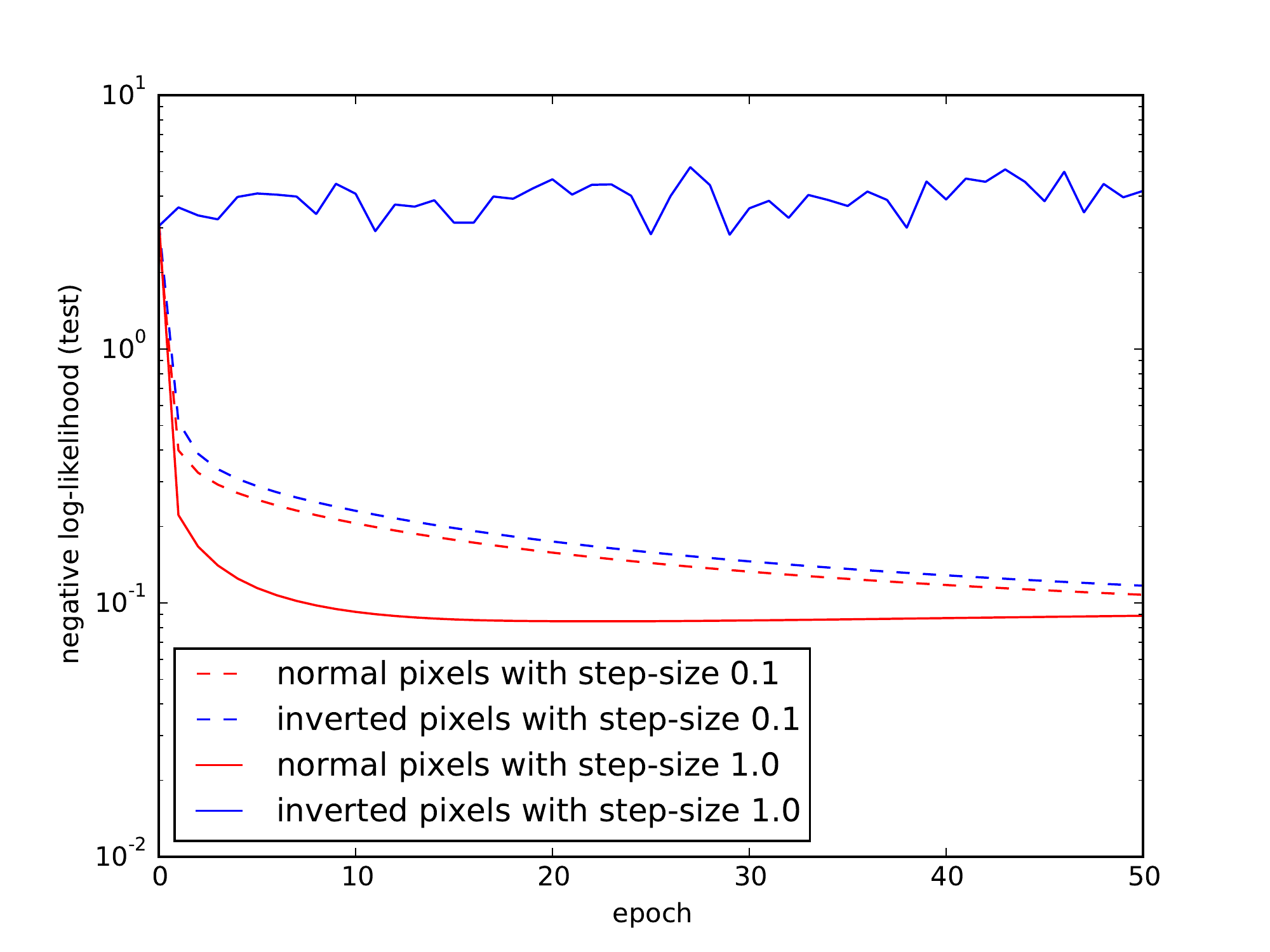}
  \caption{\label{fig:blackorwhite}Classification task on
  MNIST with a non-convolutional 784-100-10 architecture. The figure shows the impact of inverting the pixels for two different step-sizes.}
\end{figure}

Another example of lack of invariance would be a change of input
encoding. For example, if a network receives an image of a handwritten
character in black and white or in white and black, or if the inputs are
encoded with $0$ for black and $1$ for white or the other way round, this
is equivalent to changing the inputs $x$ to $1-x$. A sigmoid unit with
bias $b$ and weights $w_i$ reacts to $x$ the same way as a sigmoid unit
reacts to $1-x$ if it has bias $b'=b+\sum w_i$ and $w'_i=-w_i$
\label{blackorwhite} (indeed,
$b'+\sum w'_i(1-x_i)=b+\sum w_i x_i$). However, the gradient descent on
$(b',w')$ will behave differently from the gradient descent on $(b,w)$,
because the correspondence mixes the weights with the bias. This leads to
potentially different performances depending on whether the inputs are
white-on-black or black-on-white, as shown on \Cref{fig:blackorwhite}. This specific problem is solved by
using a tanh activation instead of a sigmoid, but it would be nice to
have optimization procedures insensitive to such simple changes.

Mathematically, the problem is that the gradient descent
\eqref{eq:euclgrad} for a differentiable function $f \from E \to \R$
defined on an abstract vector space $E$, is only well-defined after a
choice of (orthonormal) basis.  Indeed, at each point $\theta$ of $E$, the
differential $\frac{\partial f}{\partial \theta}$ is a linear form (row
vector) taking a vector $v$ as an argument and returning a scalar,
$\frac{\partial f}{\partial
\theta}\cdot v$, the
derivative of $f$ in direction $v$ at the current point $\theta$.  This
differential is converted into a (row) vector by the definition of the
gradient: $\nabla\! f$ is the unique vector such that 
$\frac{\partial f}{\partial
\theta}\cdot v = \langle \nabla\! f, v \rangle$ for all vectors $v \in E$. This
clearly depends on the definition of the inner product $\langle \cdot,
\cdot \rangle$.

In an orthonormal basis, we simply have
$\nabla\!f=\transp{(\frac{\partial f}{\partial
\theta})}$. In a non-orthonormal basis, we have
$\nabla\!f=M^{-1}\transp{(\frac{\partial f}{\partial
\theta})}$ where $M$ is the symmetric, positive-definite matrix defining
the inner product in this basis.

Using the vanilla gradient descent \eqref{eq:euclgrad} amounts to deciding
that whatever basis for $\theta$ we are currently working in, this basis
is orthonormal.  For two different parameterizations of the same intrinsic
quantity, the vanilla gradient descent produces two completely different
trajectories, and may eventually reach different local minima.
As a striking example, if we consider all points obtained with one
gradient step by varying the inner product, we span the whole half-space
where $\frac{\partial f}{\partial \theta}\cdot v > 0$.

Notice that an inner product defines a norm $\norm{v}^2\deq\langle
v,v\rangle$, which gives the notion of
length between the points of $E$. The choice of the inner product makes
it easier or more difficult for the gradient descent to move in certain
directions. Indeed the gradient descent \eqref{eq:euclgrad} is equivalent to
\begin{equation}
\theta \gets \theta+\argmin_{\delta\theta} \{ f(\theta
+\delta\theta)+\norm{\delta\theta}^2/2\eta\}
\end{equation}
up to $O(\eta^2)$ for small learning rates $\eta$: this is a minimization
over $f$, penalized by the norm of the update. Thus the
choice of norm $\norm{\cdot}$ clearly influences the direction of the update.

In general, it is not easy to define a relevant choice of inner product
$\langle\cdot, \cdot \rangle$ if nothing is known about the problem.
However, in statistical learning, it is possible to find a canonical
choice of inner product that \emph{depends on the current point
$\theta$}. At each point $\theta$, the inner product between two vectors
$v$, $v'$ will be $\langle v, v' \rangle_\theta=\transp{v}M(\theta)v'$,
where $M(\theta)$ is a particular positive definite matrix depending on
$\theta$.

Several choices for $\langle v, v' \rangle_\theta$ are given below, with the property that
the scalar product $\langle v, v' \rangle_\theta$, and the associated
\emph{metric} $\norm{v}_\theta$, do not depend on a
choice of basis for the vector space $E$. (Note that the matrix $M(\theta)$
\emph{does} depend on the basis, since the expression of $v$ and $v'$ as
vectors does.) The resulting \emph{Riemannian} gradient trajectory
\begin{equation}
\label{eq:riemgrad}
\theta \leftarrow \theta - \eta \,M(\theta)^{-1}\transp{\frac{\partial
f}{\partial \theta}} \qquad
\text{(Riemannian gradient descent)}
\end{equation}
is thus invariant to a change of basis in $E$. Actually, when the
learning rate $\eta$ tends to $0$, the resulting continuous-time
trajectory is invariant to \emph{any} smooth homeomorphism of the space
$E$, not only linear ones: this trajectory is defined on $E$ seen as a
``manifold''.


Thus, the main idea behind invariance to coordinate changes is to define an
inner product that does not depend on the numerical values used to
represent the parameter $\theta$. We now give several such constructions
together with the associated gradient update involving $M(\theta)^{-1}$.

%

\paragraph{Invariant gradient descents for neural networks.}
In machine learning, we have a training dataset $\data = (x_n,
t_n)_{n=1}^N$ where $x_n$ is a datum and $t_n$ is the associated target
or label; we will denote $\data_x$ the set of all $x_n$'s in $\data$, and likewise
for $\data_t$.
Suppose that we have a model, such as a neural network, that for each input
$x$ produces an output $y=y(x,\theta)$ depending on some parameter
$\theta$ to be trained.

We suppose that the outputs $y$ are
interpreted as a probability distribution over the possible targets $t$.
For instance, in a classification task $y$ will be a vector containing
the probabilities of the various possible labels. In a regression task,
we might assume that the actual target follows a normal distribution
centered at the predicted value $y$, namely $t_n=y(x_n,\theta)+\sigma
\Norm(0,\Id)$ (in that case, $\sigma$ may be considered an additional
parameter).

Let $p(t|y)$ be the probability distribution on the targets $t$ defined
by the output $y$ of the network. We consider the log-loss function for
input $x$ and target $t$:
\begin{equation}
\ell(t,x)\deq -\ln p(t|y(x,\theta))
\end{equation}
For instance, with a Gaussian model $t=y(x,\theta)+\sigma
\Norm(0,\Id)$, this log-loss function is
$\frac{(y-t)^2}{2\sigma^2}+\frac{\dim(y)}{2} \ln (2\pi \sigma^2)$,
namely, the square error $(y-t)^2$ up to a constant.

In this situation, there is a well-known choice of invariant scalar product on
parameter space $\theta$, for which the matrix $M(\theta)$ is the
\emph{Fisher information matrix} on $\theta$,
\begin{equation}
M^{nat}(\theta) \deq \hat{\E}_{x \in \data_{x}} \E_{\tilde{t} \sim
p(\cdot|y)} \left[ \partial_\theta \ell(\tilde t,x) \transp{(\partial_\theta
\ell(\tilde t,x))} \right] \label{eq:fish}
\end{equation}
where $\hat{\E}_{x \in \data_x}$ is the empirical average over
the feature vectors of the dataset, and where $\partial_\theta
\ell(\tilde t,x)$ denotes the (column) vector of the derivatives of the
loss with respect to the parameter $\theta$. For neural networks the
latter is computed by backpropagation. Note that this expression involves
the losses on all possible values of the targets $\tilde t$, not only the
actual target for each data sample.

The \emph{natural gradient} is the corresponding gradient descent \eqref{eq:riemgrad}, with $f\deq
\E_{(x,t)\in \data} \ell(t,x)$ the
average loss over the dataset:
\begin{equation}
\theta\gets \theta-\eta\, M^{nat}(\theta)^{-1}\,\partial_\theta f
\end{equation}

However, the natural gradient has several features that make it
unsuitable for most large-scale learning tasks.
Most of the literature on natural gradients for neural networks since
Amari's work \cite{Amari:1998} deals
with these issues.

First, the Fisher matrix is a full matrix of size $\dim(\theta)\times
\dim(\theta)$, which makes it costly to invert if not impossible to
store. Instead we will use a low-storage, easily inverted
\emph{quasi-diagonal} version of these matrices, which keeps many
invariance properties of the full matrix.

Second, the Fisher matrix involves an expectation over ``pseudo-targets''
$\tilde t\sim p(\cdot|y)$ drawn from the distribution defined by the
output of the network. This might not be a problem for classification
tasks for which the number of possibilities for $t$ is small, but
requires another approach for a Gaussian model if we want to avoid
numerical integration over $\tilde t\sim y+\sigma \Norm(0,\Id)$. We
describe three ways around this: the \emph{outer product} approximation,
a Monte Carlo approximation, or an exact version which requires $\dim(y)$
backpropagations per sample.

Third, the Fisher matrix for a given value of the parameter $\theta$ is a
sum over $x$ in the whole dataset. Thus, each natural gradient update
would need a whole sweep over the dataset to compute the Fisher matrix
for the current parameter. We will use a moving average over $x$ instead.

Let us discuss each of these points in turn.

\paragraph{Quasi-diagonal Riemannian metrics.} Instead of the full Fisher
matrix, we
use the \emph{quasi-diagonal reduction} of $M^{-1}$, which involves
computing and storing only the diagonal terms and a few off-diagonal terms of the
matrix $M$.
The quasi-diagonal reduction was introduced in \cite{Ollivier2015} to
exactly keep some of the invariance properties of the Fisher matrix, at a price
close to that of diagonal matrices.

Quasi-diagonal reduction uses a decomposition of the parameter $\theta$
into blocks. For neural networks, these blocks will be the parameters
(bias and weights) incoming to each neuron, with the bias being the first
parameter in each block. In each block, only the diagonal and the first
row of the block are stored. Thus the number of non-zero entries to be
stored is $2\dim(\theta)$ (actually slightly less, since in each block
the first entry lies both on the diagonal and on the first row; in what
follows, we include it in the diagonal and always ignore the first entry
of the row).

Maintaining the first row in each block accounts for possible correlations
between biases and weights, such as those appearing when the input values
are transformed from $x$ to $1-x$ (see p.~\pageref{blackorwhite}).  This
is why the bias plays a special role here and has to be the first
parameter in each block.

Quasi-diagonal metrics guarantee exact invariance to affine changes in the
activations of each unit \cite[Section 2.3]{Ollivier2015}, including
each input unit.

The operations we will need to perform on the matrix $M$ are of two types:
computing $M$ by adding rank-one contributions of the form $v\transp{v}$
with $v=\partial_\theta \ell(\tilde t,x)$ in \eqref{eq:fish},
and applying the inverse of $M$ to a vector $v$ in the parameter update
\eqref{eq:riemgrad}.
For quasi-diagonal matrices these operations are explicited in Algorithms~\ref{alg:qdacc}
and \ref{alg:qd}. The cost is about twice that of using diagonal
matrices.


A by-product of this block-wise quasi-diagonal structure is that each
layer can be considered independently.
This enables to implement the computation of the metric in a
modular fashion.

\begin{center}
\begin{algorithm}[H]
  \SetAlgoLined
\SetKw{func}{Function}
\SetKwFunction{qdsolve}{QDSolve}
\func{\qdsolve{$M$,$v$}}\\
  \KwData{Vector $v$; block decomposition for the components of $v$;
  matrix $M$ of size $\dim(v)\times \dim(v)$ of which only the diagonal
  and the first row in each block are known; regularization threshold
  $\eps\geq 0$.}
  \KwResult{Quasi-diagonal inverse $\QD(M)^{-1}.v$}
  \ForEach{block of components of $v$}{
    $w\gets$ $k$-th block of $v$\;
    $\Delta\gets$ diagonal of the $k$-th block of $M$\;
    $r\gets$ first row of the $k$-th block of $M$\;
    $\Delta \gets \Delta + \epsilon$\;
    Index the components of the block from $0$ to $n_k-1$\;
    \For{i = 1 to $n_k-1$ {\small(but NOT $i=0$})}{
      $w_i \gets \frac{\Delta_0 w_i - r_i w_0}{\max\left(\Delta_i
      \Delta_0-r_i^2,\,\epsilon\right)}$\;
    }
    $w_0\gets \frac{1}{\Delta_0}(w_0-\sum_{i=1}^{n_k-1} r_i w_i)$\;
    Store $w$ into the $k$-th block of the result\;
  }
  \caption{Pseudo-code for quasi-diagonal inversion.\label{alg:qd} The
  case when the first rows are ignored ($q=0$) corresponds to a diagonal
  inversion $\diag(M+\eps)^{-1}.v$. In our experiments, $\epsilon$ is set
  to $10^{-8}$.}
\end{algorithm}
\end{center}

\begin{center}
\begin{algorithm}[H]
  \SetAlgoLined
\SetKw{func}{Function}
\SetKwFunction{qdrankoneupdate}{QDRankOneUpdate}
\func{\qdrankoneupdate{$M$,$v$,$\alpha$}}\\
  \KwData{Vector $v$; block decomposition for the components of $v$;
  matrix $M$ of size $\dim(v)\times \dim(v)$ of which only the diagonal
  and the first row in each block are known; real number $\alpha$.}
  \KwResult{Rank-one update $M\gets M+\alpha.\QD(v\transp{v})$}
  \ForEach{block of components of $v$}{
    $w\gets$ $k$-th block of $v$\;
    $\Delta\gets$ diagonal of the $k$-th block of $M$\;
    $r\gets$ first row of the $k$-th block of $M$\;
    $\Delta \gets \Delta+ \alpha \,w^{\odot 2}$\;
    $r\gets r+\alpha\, w_0\, \transp{w}$ with $w_0$ the first entry of $w$\;
    Diagonal of the $k$-th block of $M$ $\gets \Delta$\;
    First row of the $k$-th block of $M$ $\gets r$\;
  }
  \caption{Pseudo-code for quasi-diagonal accumulation: rank-one update
  of a quasi-diagonal matrix.\label{alg:qdacc}}
\end{algorithm}
\end{center}

\paragraph{Online Riemannian gradient descent, and metric initialization.} As the Fisher matrix for a given value of the parameter $\theta$ is a
sum over $x$ in the whole dataset, each natural gradient update
would need a whole sweep over the dataset to compute the Fisher matrix
for the current parameter. This is suitable for batch learning, but not
for online stochastic gradient descent.

This can be alleviated by updating the metric
via a moving average
\begin{equation}
M \gets (1-\gamma) M + \gamma M_{\text{minibatch}}
\end{equation}
where $\gamma$ is the metric update rate and where $M_{\text{minibatch}}$
is the metric computed on a small subset of the data, i.e., by replacing
the empirical average $\hat \E_{x\in\data_x}$ with an empirical average
$\hat \E_{x\in \data'_x}$ over a minibatch $\data'\subset \data$. Typically
$\data'$ is the same
subset on which the gradient of the loss is computed in a stochastic
gradient scheme. (This subset may be reduced to one sample.) This results
in an \emph{online Riemannian gradient descent}.

In practice,
choosing $\gamma \approx \frac{1}{\# \text{minibatches}}$ ensures that the metric
is mostly renewed after one whole sweep over the dataset.
Usually we initialize the metric on the first minibatch
($\gamma=1$ for the first iteration), which is
empirically better than using the identity metric for the first update.
At startup when using very small minibatches (e.g., minibatches of size $1$) it may
be advisable to initialize the metric on a larger number of samples
before the first parameter update. (An alternative is to initialize the
metric to $\Id$, but this breaks invariance at startup.)

\paragraph{Natural gradient, outer product, and Monte Carlo natural
gradient.}
The Fisher matrix involves an expectation over ``pseudo-targets''
$\tilde t\sim p(\cdot|y)$ drawn from the distribution defined by the
output of the network: a backpropagation is needed for each possible
value of $\tilde t$ in order to compute $\partial_\theta \ell(\tilde t,x)$. This is acceptable only for classification
tasks for which the number of possibilities for $t$ is small.
Let us now describe three ways around this issue.

A first way to avoid the expectation over pseudo-targets $\tilde{t}\sim
p(\cdot|y)$ is to only use the actual targets in the dataset. This
defines the \emph{outer product} approximation of the Fisher matrix
%
\begin{equation}
M^{OP}(\theta) \deq \hat{\E}_{(x,t) \in \data}
\left[ \partial_\theta \ell(\tilde t,x) \transp{(\partial_\theta
\ell(\tilde t,x))} \right] \label{eq:OP}
\end{equation}
thus replacing $\hat{\E}_{x \in \data_{x}} \E_{\tilde{t} \sim p(\cdot|y)}$ with the empirical average $\hat{\E}_{(x,t) \in \data}$.
For each training sample, we have to compute a rank-one matrix given by the
outer product of the gradient for this sample, hence the name.
This method has sometimes been used directly under the name ``natural
gradient'', although it has different properties, see discussion in
\cite{pascanu:2013} and \cite{Ollivier2015}.

An
advantage is that it can be computed directly from the gradient provided by
usual backpropagation on each sample. The corresponding pseudocode is
given in Algorithm~\ref{alg:qdop} for a minibatch of size $1$.

\begin{center}
\begin{algorithm}[H]
  \SetAlgoLined
  \KwData{Dataset $\data$, a neural network structure with parameters
  $\theta$}
  \KwResult{optimized parameters $\theta$}
  \While{not finish}{
      retrieve a data sample $x$ and corresponding target $t$ from $\data$\;
      forward $x$ through the network\;
      compute loss $\ell(t,x)$\;
      backpropagate and compute derivative of loss: $v\gets
      \partial_\theta \ell(t,x)$\;
      update quasi-diagonal metric using $v\transp{v}$:\\
      \Indp
      $M\gets (1-\gamma)M$\;
      \qdrankoneupdate{$M$,$v$,$\gamma$}\;
      \Indm
      apply inverse metric: $v\gets$ \qdsolve{$M$,$v$}\;
      update parameters: $\theta\gets \theta-\eta v$\;
  }
  \caption{Online gradient descent using the quasi-diagonal outer product
  metric.\label{alg:qdop}}
\end{algorithm}
\end{center}

A second way to proceed is to replace the expectation over pseudo-targets
$\tilde t$ with a Monte Carlo approximation using $n_{MC}$ samples:
\begin{equation}
M^{MCnat}(\theta)\deq \E_{x\in \data_x} \left[ \frac{1}{n_{MC}}
\sum_{i=1}^{n_{MC}} 
\partial_\theta \ell(\tilde t_i,x) \transp{(\partial_\theta
\ell(\tilde t_i,x))} \right] \label{eq:MCnat}
\end{equation}
where each pseudo-target $\tilde t_i$ is drawn from the distribution
$p(\tilde t|y)$ defined by the output of the network for each input $x$.
This is the \emph{Monte Carlo natural gradient} \cite{Ollivier2015}. We
have found that $n_{MC}=1$ (one pseudo-target for each input $x$) works well in practice.

The corresponding pseudocode is
given in Algorithm~\ref{alg:qdmcnat} for a minibatch of size $1$ and
$n_{MC}=1$.

\begin{center}
\begin{algorithm}[H]
  \SetAlgoLined
  \KwData{Dataset $\data$, a neural network structure with parameters
  $\theta$}
  \KwResult{optimized parameters $\theta$}
  \While{not finish}{
      retrieve a data sample $x$ and corresponding target $t$ from $\data$\;
      forward $x$ through the network\;
      compute loss $\ell(t,x)$\;
      backpropagate and compute derivative of loss: $v\gets
      \partial_\theta \ell(t,x)$\;
      generate pseudo-target $\tilde t$ according to probability
      distribution defined by the output layer $y$ of the network:
      $\tilde t\sim p(\cdot|y)$\;
      backpropagate and compute derivative of loss for $\tilde t$:
      $\tilde v\gets \partial_\theta \ell(\tilde t,x)$\;
      update quasi-diagonal metric using $\tilde v\transp{\tilde v}$:\\
      \Indp
        $M\gets (1-\gamma)M$\;
        \qdrankoneupdate{$M$,$\tilde v$,$\gamma$}\;
      \Indm
      apply inverse metric: $v\gets$ \qdsolve{$M$,$v$}\;
      update parameters: $\theta\gets \theta-\eta v$\;
  }
  \caption{Online gradient descent using the quasi-diagonal Monte Carlo
  natural gradient, with $n_{MC}=1$.
  \label{alg:qdmcnat}}
\end{algorithm}
\end{center}

The third option is to compute the expectation over $\tilde{t}$ exactly
using algebraic properties of the Fisher matrix. This can be done at the
cost of $\dim(y)$ backpropagations per sample instead of one.
The details depend on the type of the output layer, as follows.

If the
output is a softmax over $K$ categories, one can just write out the
expectation explicitly:
\begin{equation}
M^{nat}(\theta) = \hat{\E}_{x \in \data_{x}} \left[ \sum_{\tilde{t}=1}^K
p(\tilde t|y) \,\partial_\theta \ell(\tilde t,x) \transp{(\partial_\theta
\ell(\tilde t,x))} \right]
\end{equation}
so that the metric is a sum of rank-one terms over all possible
pseudo-targets $\tilde t$, weighted by their predicted probabilities.

If the output model is multivariate Gaussian with diagonal covariance matrix
$\Sigma=\diag(\sigma_k^2)$ ($\Sigma$ may be known or learned), i.e.,
$t=y+\Norm(0,\Sigma)$, then one can prove that the Fisher matrix is equal to
\begin{equation}
M^{nat}(\theta) = \hat{\E}_{x \in \data_{x}} \left[ \sum_{k=1}^K
\frac{1}{\sigma_k^2} \,\partial_\theta y_k
\transp{(\partial_\theta y_k)} \right]
\end{equation}
where $y_k$ is the $k$-th component of the network output. The derivative
$\partial_\theta y_k$ can be obtained by backpropagation if the
backpropagation is initialized by setting the $k$-th output unit to $1$
and all other units to $0$. This has to be done separately for each
output unit $k$. Thus, for each input $x$ the metric is a contribution of
several rank-one terms $\partial_\theta y_k
\transp{(\partial_\theta y_k)}$ obtained by backpropagation, and weighted
by $1/\sigma_k^2$.

Another model for predicting binary data is the Bernoulli output, in which the
activities of output units $y_k\in [0;1]$ are interpreted as probabilities to
have a $0$ or a $1$. This case is similar to the Gaussian case up to
replacing $\frac{1}{\sigma_k^2}$ with
$\frac{1}{y_k(1-y_k)}$ (inverse variance of the Bernoulli variable defined by
output unit $k$).

The pseudocode for the quasi-diagonal natural gradient is given in
Algorithm~\ref{alg:qdnat}.

\begin{center}
\begin{algorithm}[H]
  \SetAlgoLined
  \KwData{Dataset $\data$, a neural network structure with parameters
    $\theta$, a quasi-diagonal metric $M$}
  \KwResult{optimized parameters $\theta$}
  \While{not finish}{
      retrieve a data sample $x$ and corresponding target $t$ from $\data$\;
      forward $x$ through the network\;
      compute loss $\ell(t,x)$\;
      backpropagate and compute derivative of loss: $v\gets
      \partial_\theta \ell(t,x)$\;
      $M\gets (1-\gamma)M$\;
      Depending on output layer interpretation:\\
      \textit{Categorical output with classes \{1,\ldots,K\}}:\\
      \Indp
      \For{each class $\tilde t$ from $1$ to $K$}{
        Set $\tilde t$ as the pseudo-target\;
	$\alpha\gets p(\tilde t|y)$ \;
        backpropagate and compute derivative of loss for $\tilde t$:
        $\tilde v\gets \partial_\theta \ell(\tilde t,x)$\;
        update quasi-diagonal metric using $\tilde v\transp{\tilde v}$:\\
          \qdrankoneupdate{$M$,$\tilde v$,$\gamma \alpha$}\;
      }
      \Indm
      \textit{Gaussian output (square loss) with variance $\diag(\sigma_k^2)$, $k=1,\ldots,K$}:\\
      \Indp
      \For{each output unit $k$ from $1$ to $K$}{
	$\alpha\gets 1/\sigma_k^2$ \;
        $\tilde v\gets \partial_\theta y^k$ (obtained by 
	setting backpropagated values on the $k$-th output unit to $1$, all
	the others to $0$, and backpropagating)\;
        update quasi-diagonal metric using $\tilde v\transp{\tilde v}$:\\
          \qdrankoneupdate{$M$,$\tilde v$,$\gamma \alpha$}\;
      }
      \Indm
      apply inverse metric: $v\gets$ \qdsolve{$M$,$v$}\;
      update parameters: $\theta\gets \theta-\eta v$\;
  }
  \caption{Online gradient descent using the quasi-diagonal
  natural gradient.
  \label{alg:qdnat}}
\end{algorithm}
\end{center}

Notably, the OP and Monte Carlo approximation both keep the invariance
properties of the natural gradient. This is not the case for other
natural gradient approximations such as the blockwise rank-one
approximation used in \cite{TONGA}.

Each of these approaches has its strengths and weaknesses. The
(quasi-diagonal) OP is easy to implement as it relies only on quantities
(the gradient for each sample) that have been computed anyway, while the
Monte Carlo and exact natural gradient must make backpropagation passes
for other values of the output of the network.

When the model fits the data well, the distribution $p(\tilde
t|y)$ gives high probability
to the actual targets $t$ and thus, the OP metric is a good approximation of the Fisher metric.
However, at startup, the model does not fit the data and  OP might
poorly approximate the Fisher metric. Similarly, if the output model is
misspecified, OP can perform badly even in the last stages of
optimization; for instance, OP fails miserably to optimize a quadratic
function in the absence of noise, an example to keep in mind.\footnote{Indeed, suppose that the loss
function is $\norm{\theta-x}^2/2\sigma^2$, corresponding to the log-loss of a
Gaussian model with variance $\sigma^2$, and that all the data points are $x=0$. Then the gradient
of the loss is $\theta/\sigma^2$, the OP matrix is the square gradient
$\theta^2/\sigma^4$, and the OP
gradient descent is $\theta \gets \theta - \eta\sigma^2/\theta$. This is much too slow
at startup and much too fast for final convergence. On the other hand,
the natural gradient is $\theta\gets \theta-\eta \theta$ which behaves
nicely. The catastrophic behavior of OP for small $\theta$ reflects the
fact that the data have variance $0$, not $\sigma^2$. Its bad behavior
for large $\theta$ reflects the fact that the data do not follow the
model $\Norm(\theta,\sigma^2)$
at startup. In both cases, the OP approximation is unjustified
hence a huge difference
with the natural gradient. The Monte Carlo
natural gradient will behave better in this instance. This particular
divergence of the OP gradient for $\theta$ close to $0$ disappears as
soon as there is some noise in the data.}

The exact quasi-diagonal natural gradient is only affordable when the
dimension of the output $y$ of the network is not too large, as $\dim(y)$
backpropagations per sample are required. Thus the OP and Monte Carlo approximations
are appealing when output dimensionality is large, as is typical for
auto-encoders. However, in this case 
the probability distribution $p(\tilde t|y)$ is defined over a
high-dimensional space, and the OP approximation using the single
deterministic point $\tilde
t=t$ may be poor. (In preliminary experiments from \cite{Ollivier2015}, the
OP approximation performed poorly for an auto-encoding task.) On the
other hand, Monte Carlo integration often performs relatively well in
high dimension, and might be a sensible choice for large-dimensional
outputs such as in auto-encoders.

Still, the choice between these three options is largely experimental.
Since all three have closely related implementations
(Algorithms~\ref{alg:qdop}--\ref{alg:qdnat}), they can be compared with
little effort.

\paragraph{Diagonal versions: From AdaGrad to OP, and
invariance properties.}
The algorithms presented above with quasi-diagonal matrices also have a
diagonal version, obtained by simply discarding the non-diagonal terms.
(This breaks affine invariance.)

For instance, the update for diagonal OP (DOP) reads
\begin{equation}
M^{DOP} \gets (1-\gamma)M^{DOP}+\gamma \,\diag(\partial_\theta \ell
\,\transp{\partial_\theta \ell})
\end{equation}
where $\partial_\theta \ell$ is the derivative of the loss $\ell$ for the
current sample.
This can be rewritten on the vector of diagonal entries of $M^{DOP}$ as
\begin{equation}
\diag(M^{DOP}) \gets (1-\gamma)\diag(M^{DOP})+\gamma\,(\partial_\theta
\ell)^{\odot 2}
\end{equation}
This update is the same as the one used in the family of AdaGrad or
RMSProp algorithms, except that the latter use the \emph{square root} of
$M$ in the final parameter update:
\begin{align}
\theta \leftarrow \theta - \eta M_{Ada}^{-1/2} \partial_\theta \ell
\end{align}
where $M_{Ada}$ follows the same update as $M^{DOP}$.
Thus, although AdaGrad shares a similar framework with DOP, it is not
naturally interpreted as a Riemannian metric on parameter space (there is
no well-defined trajectory on the parameter space seen as a manifold), 
because taking the element-wise square-root breaks all potential invariances.

To summarize, the \emph{diagonal} versions DOP, DMCNat and DNat are exactly
invariant to \emph{rescaling} of each parameter component.  In addition, the
\emph{quasi-diagonal} versions of these algorithms are exactly invariant
to an \emph{affine change in the activity of each unit}, including the
activities of input units; this covers, for instance, using tanh instead
of sigmoid, or using white-on-black instead of black-on-white inputs. In
contrast, to our knowledge AdaGrad has no invariance properties (the
\emph{values} of the gradients $M_{Ada}^{-1/2} \partial_\theta \ell$ are scale-invariant in AdaGrad, but not
the resulting parameter trajectories, since the parameter $\theta$ is
updated by the same value whatever its scale).

\paragraph{Experimental studies.} We demonstrate the invariance and
efficiency of Riemannian gradient descents through experiments on a
series of classification and regression tasks. For each task, we choose
an architecture and an activation function, and we perform a grid search
over various powers of $10$ for the step-size. The step-size is kept
fixed during optimization.  Then, for each algorithm, we report the curve
associated with the best step size.  The algorithms tested are standard
stochastic gradient descent (SGD), AdaGrad, and the diagonal and
quasi-diagonal versions of OP, of Monte Carlo natural gradient (with only
one sample), and of the exact natural gradient when the dimension of the
output layer is not too large to compute it.

First, we study the classification task on 
MNIST \cite{mnistlecun}
and the Street View Housing Numbers (SVHN)\cite{Netzer:2011}.
We work in a permutation-invariant setting,
i.e., the network is not convolutional and the natural topology of the image is not taken into account in the network structure.
We also converted the SVHN images into grayscale images in order to reduce the dimensionality of the input.

Note that these experiments are small-scale and not geared towards obtaining
state-of-the-art performance, but aim at comparing the behavior of several
algorithms on a common ground, with relatively small
architectures.\footnote{In particular, we have not used
the test sets of these datasets, only validation sets extracted from the
training sets. This is because we did not want to use information from
the test sets (e.g., hyperparameters) before testing the Riemannian
algorithms on other architectures, which will be done in future work.}

\begin{figure}[t]
  \centering
  \includegraphics[scale=0.35]{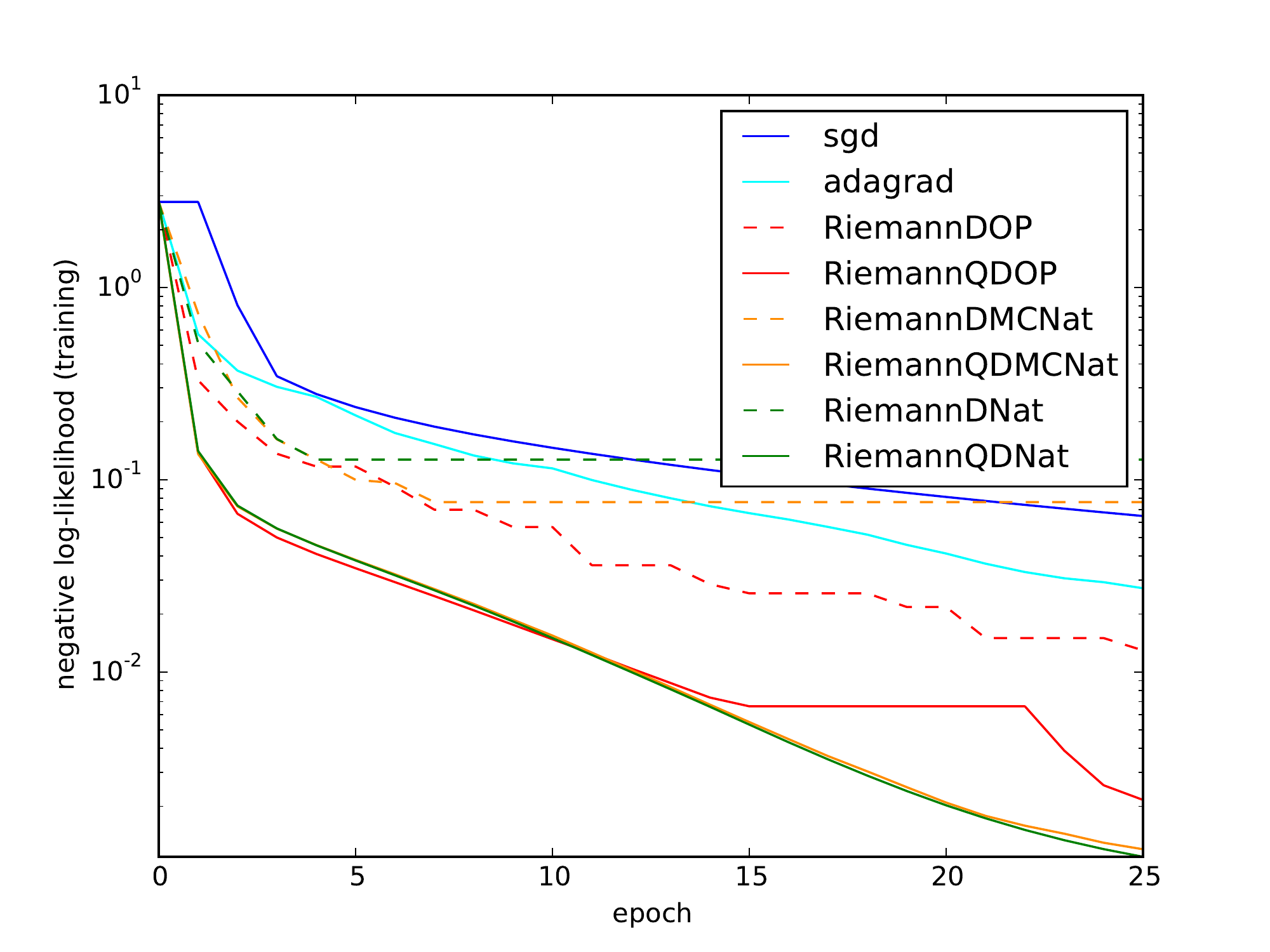}
  \includegraphics[scale=0.35]{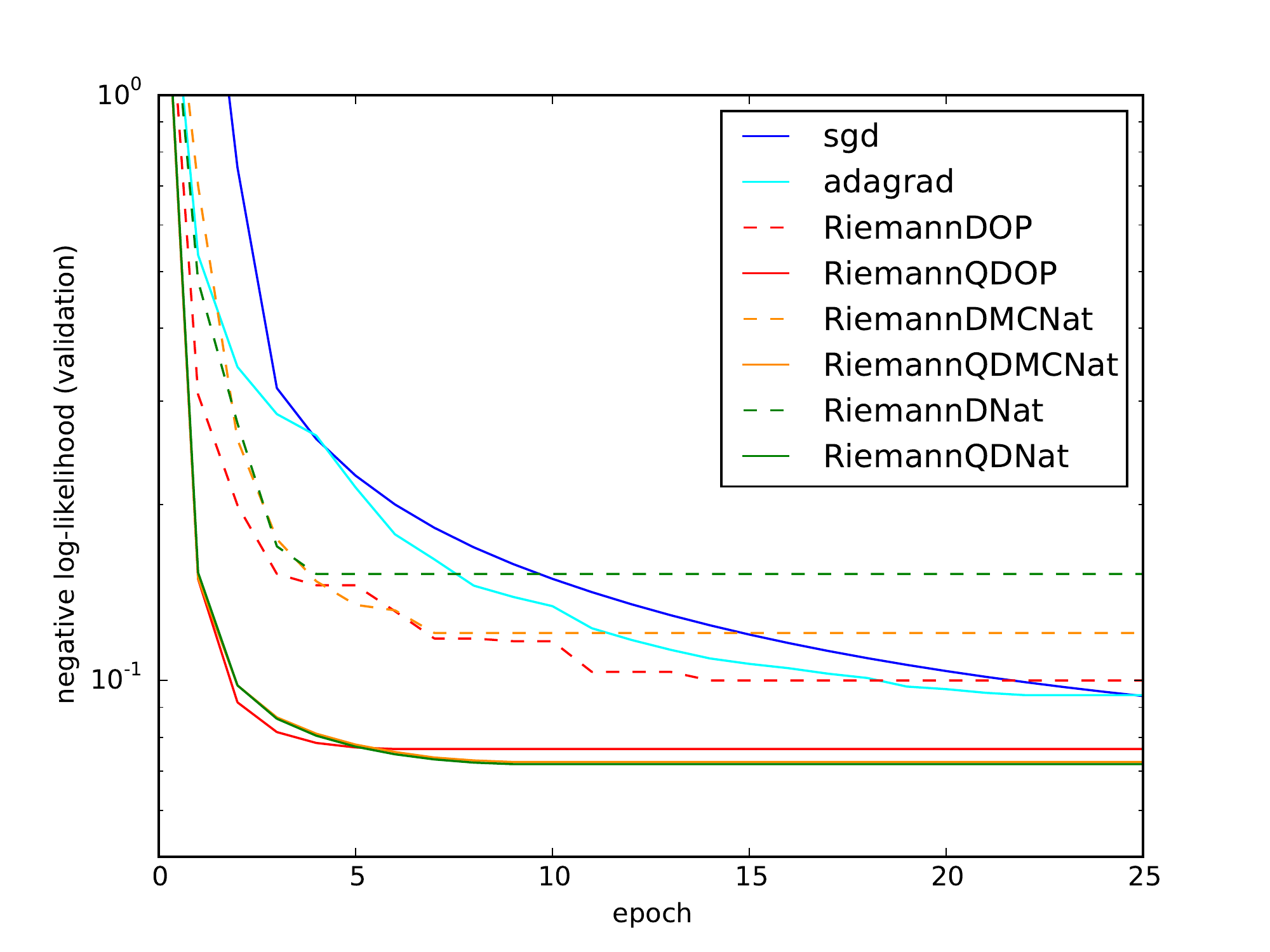}
  \includegraphics[scale=0.35]{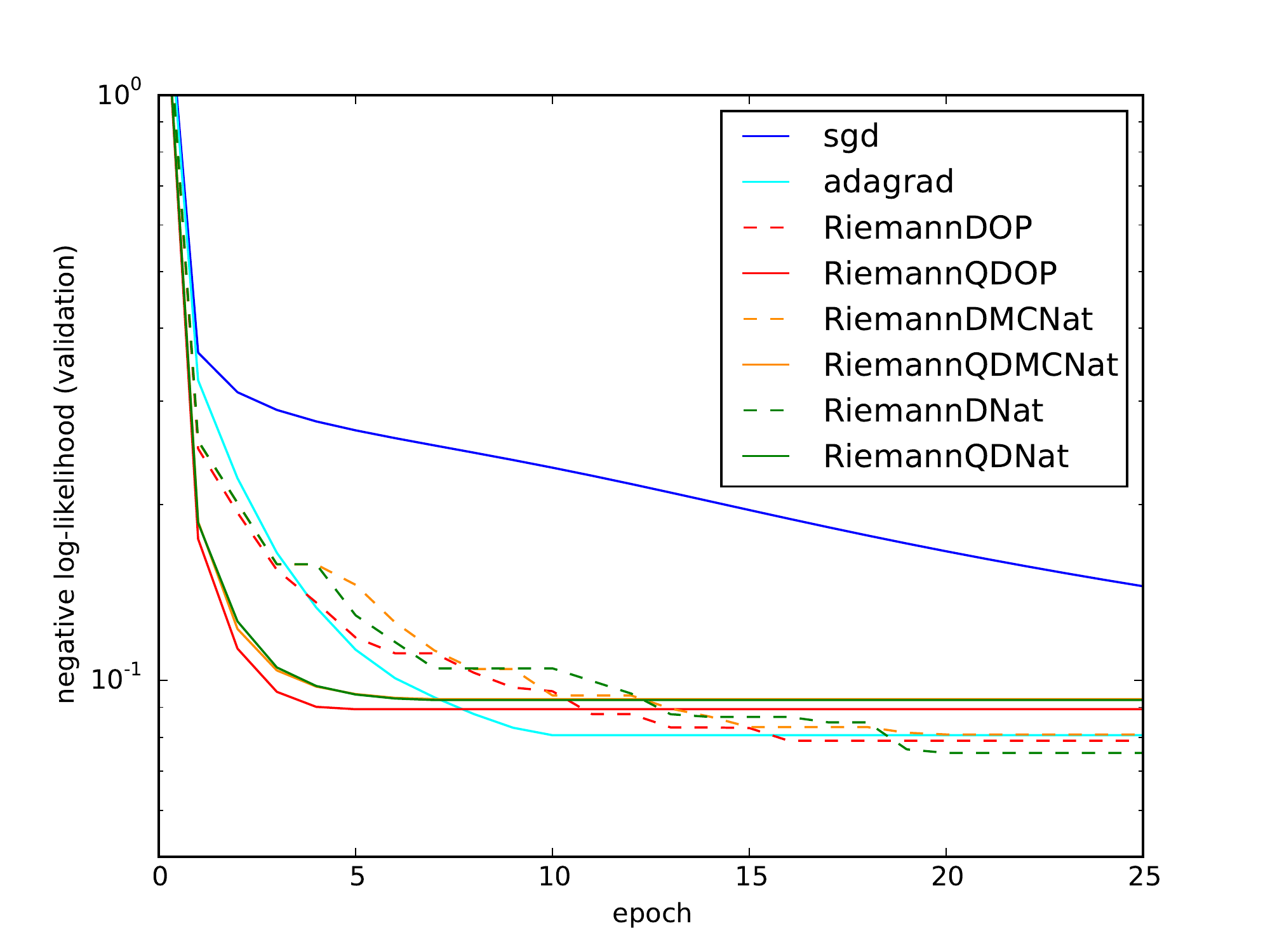}
  \includegraphics[scale=0.35]{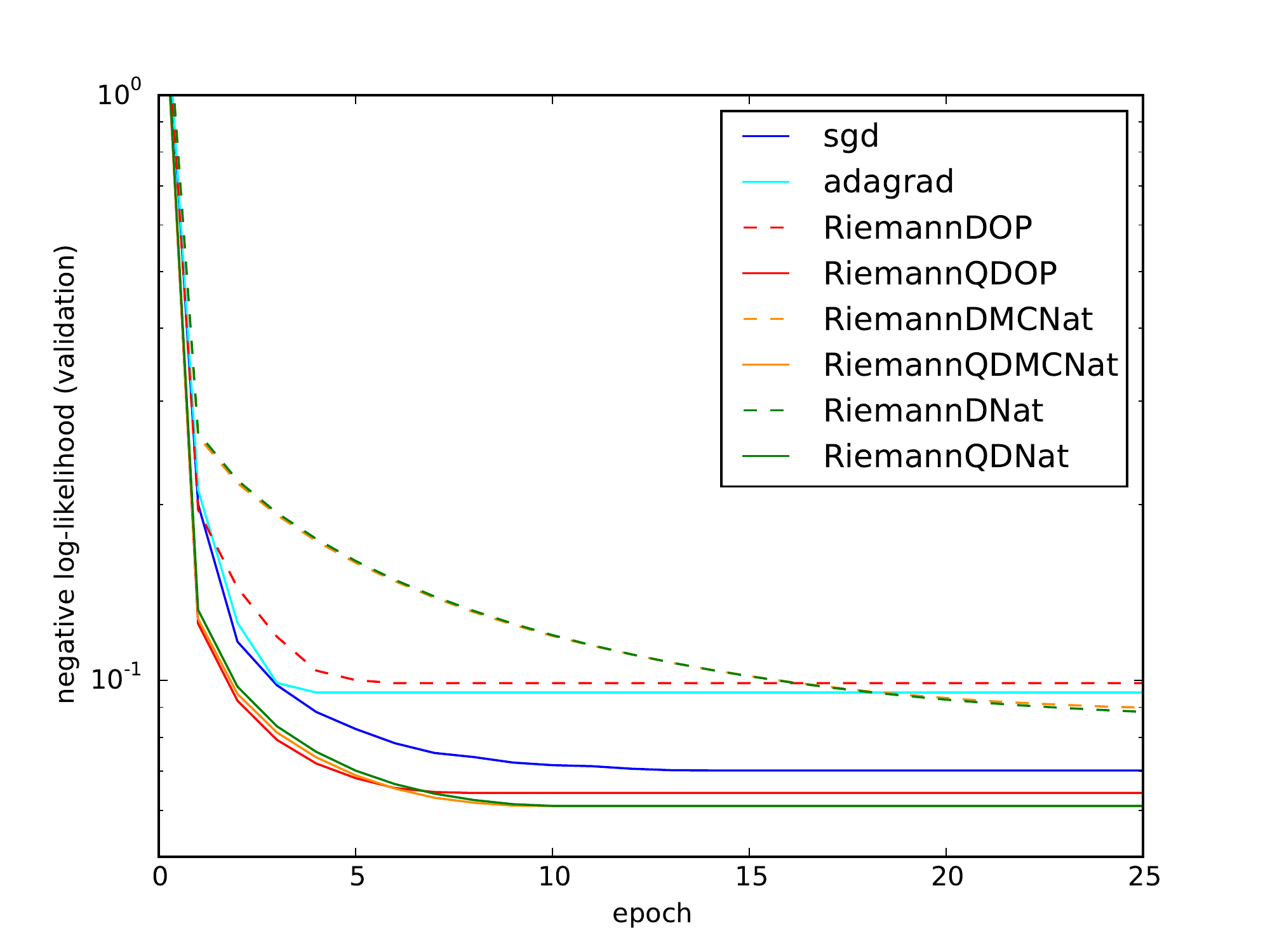}
  \caption{\label{fig:mnist_nll_NN800-800}Classification task on MNIST
  with a non-convolutional 784-800-800-10 architecture without
  regularization. We use a sigmoid (upper left and upper right), a tanh
  (lower left) and
  a ReLU (lower right).}
\end{figure}

Experiments on the classification task confirm that Riemannian algorithms
are both more invariant and more efficient than non-invariant algorithms.  In
the first experiment we train a network with two hidden layers and 800
hidden units per layer, without any regularization, on MNIST; the results
are given on \Cref{fig:mnist_nll_NN800-800}. 

Quasi-diagonal algorithms (plain lines) converge faster than the other
algorithms.  Especially, they exhibit a steep slope for the first few
epochs, and quickly reach a satisfying performance.  Their trajectories
are also very similar for every activation function.\footnote{Note that
their invariance properties guarantee a similar performance for sigmoid
and tanh as they represent equivalent models using different variables,
but not necessarily for ReLU which is a different model.}  The
trajectories of SGD, AdaGrad, and the diagonally approximated Riemannian algorithms are more
variable: for instance, SGD is close enough to the quasi-diagonal
algorithms with ReLU activation but not with sigmoid or tanh, and AdaGrad
performs well with tanh but not with sigmoid or ReLU.  
Finally, the Monte Carlo QD natural gradient seems to be a
very good approximation of the exact QD natural gradient, even with only one
Monte Carlo sample.  This may be related to simplicity
of the problem, since the probabilities of the output layer converges
rapidly to their optimal values.   

Quasi-diagonal algorithms are also efficient in practice.  Their
computational cost is reasonably close to pure SGD, as shown on
\Cref{table:computation}, with an overhead of about $2$ as could be
expected. They are often much faster to learn in terms of
number of
training examples processed, compared to SGD or AdaGrad.

\begin{figure}[t]
  \centering
  \includegraphics[scale=0.35]{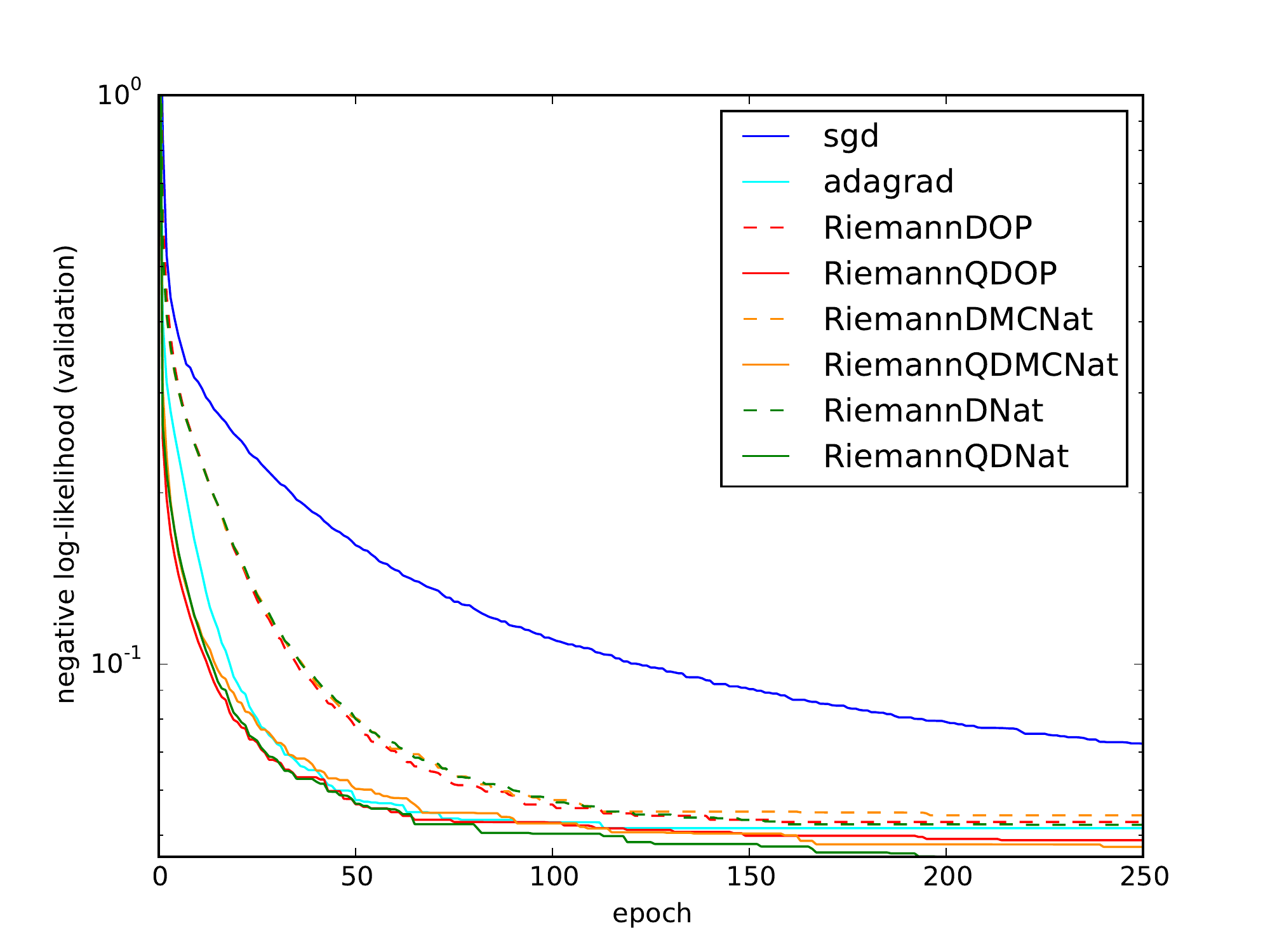}
  \includegraphics[scale=0.35]{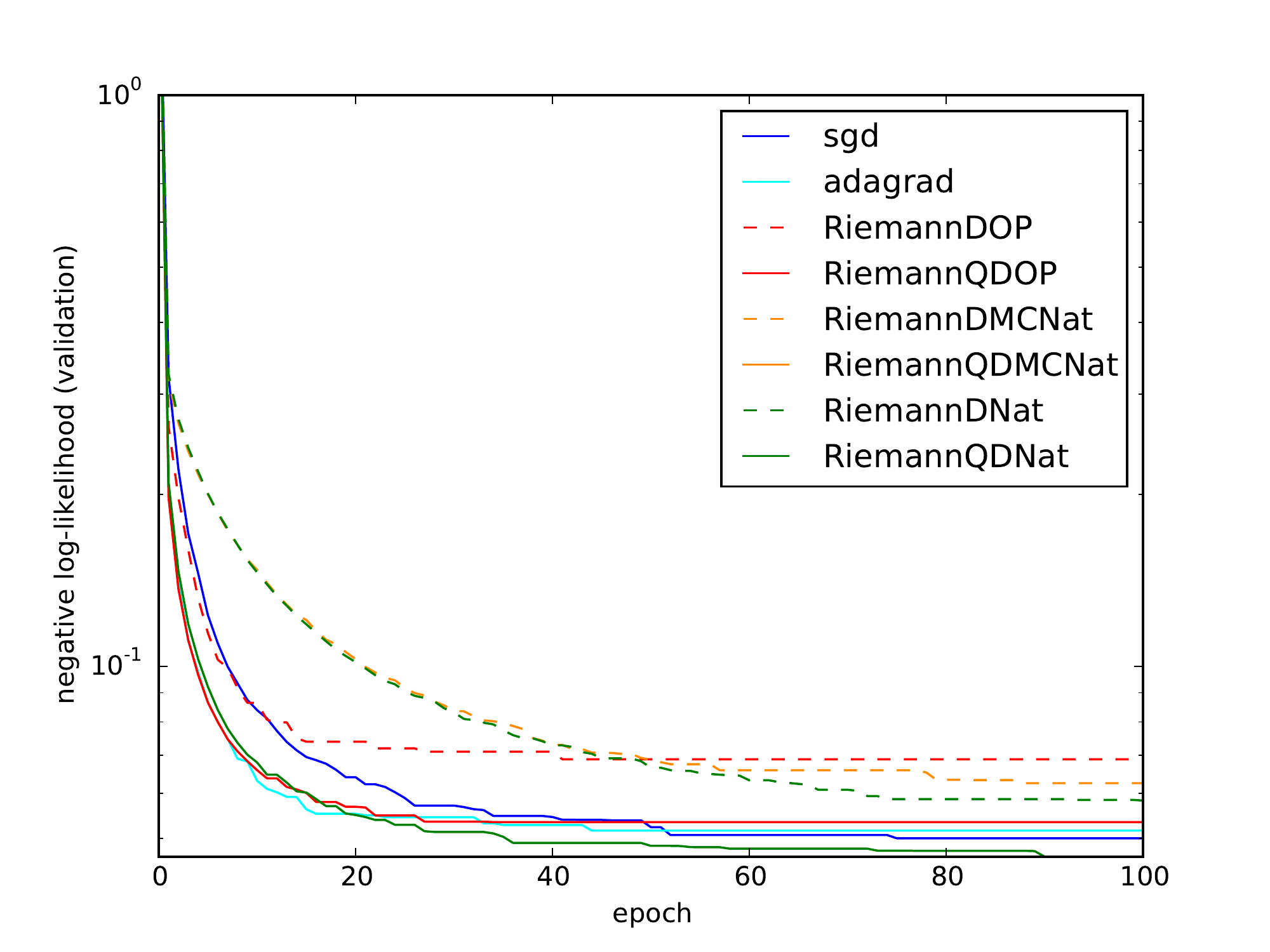}
  \caption{\label{fig:mnist_drop_nll_NN800-800}Classification task on
  MNIST with a non-convolutional 784-800-800-10 architecture and sigmoid (left) or ReLU (right). The network is regularized with dropout.}
\end{figure}

The quasi-diagonal algorithms also
behave well with the dropout regularization \cite{srivastava14a}, as can be seen
on \Cref{fig:mnist_drop_nll_NN800-800}. Several runs went below $1\%$
classification error with this simple 800-800, permutation-invariant
architecture.
Three groups of trajectories are clearly visible on
\Cref{fig:mnist_drop_nll_NN800-800}, with AdaGrad fairly close to the
quasi-diagonal algorithms on this example. There is a clear difference
between the quasi-diagonal algorithms and their diagonal approximations:
keeping only the diagonal breaks the invariance to affine transormations
of the activities, which thus appears as a key factor here as well as in
almost all experiments below.

\begin{table}[t]
  \centering
  \begin{tabular}{|l|l|}\hline
    Algorithm & mean (std) \\ \hline \hline
    SGD & 29.10 (0.76) \\
    AdaGrad & 33.39 (2.26) \\
    RiemannDOP & 42.44 (1.26) \\
    RiemannQDOP & 56.68 (3.75) \\
    RiemannDMCNat & 58.17 (4.57) \\
    RiemannQDMCNat & 68.01 (2.35) \\
    RiemannDNat & 120.16 (0.34) \\
    RiemannQDNat & 129.76 (0.34) \\\hline
  \end{tabular}
  \caption{\label{table:computation}Computational time per epoch for a
  784-800-800-10 architecture on MNIST with minibatch of size 500 on a CPU.}
\end{table}

%

\begin{figure}[t]
  \centering
  \includegraphics[scale=0.35]{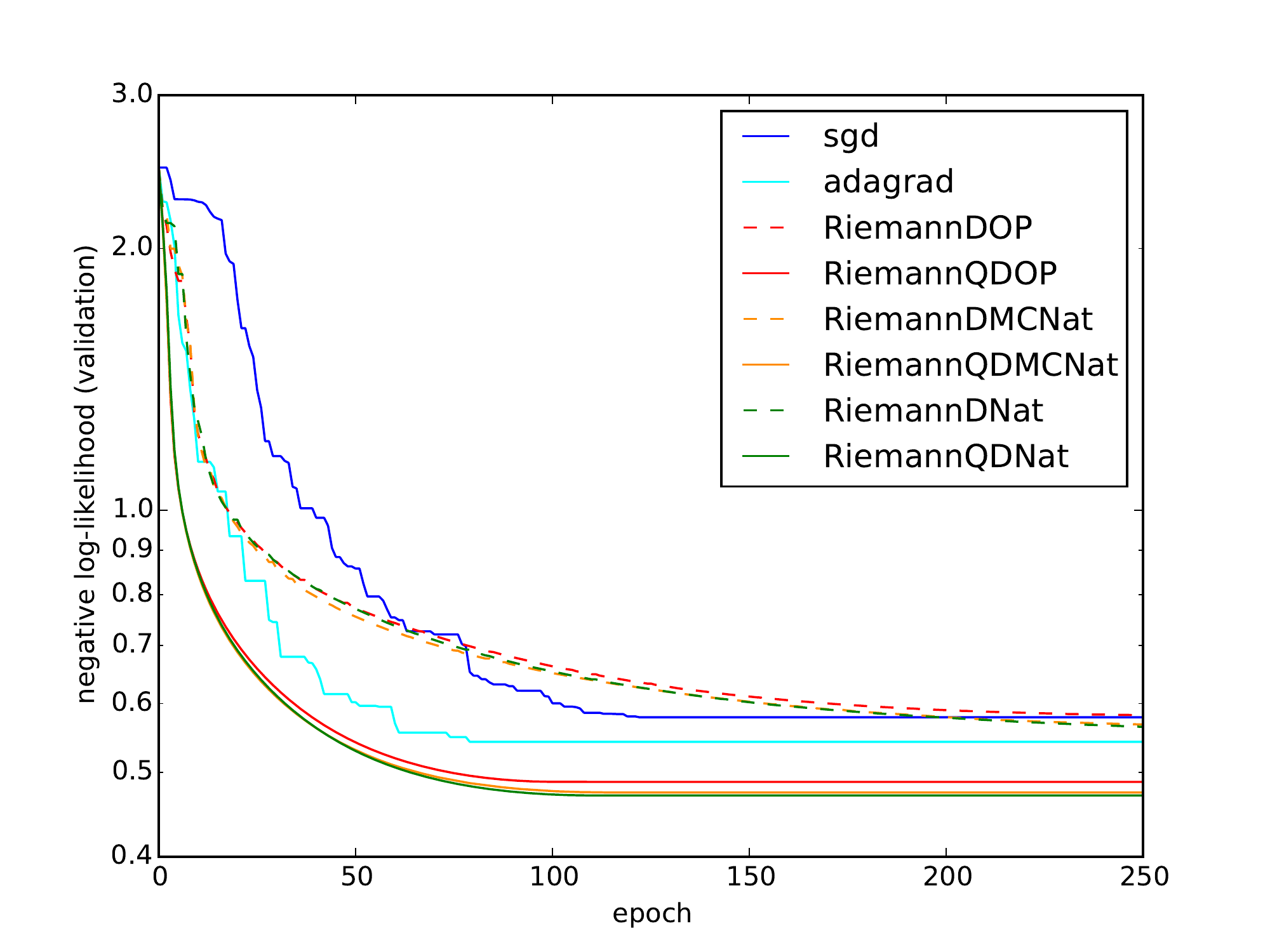}
  \includegraphics[scale=0.35]{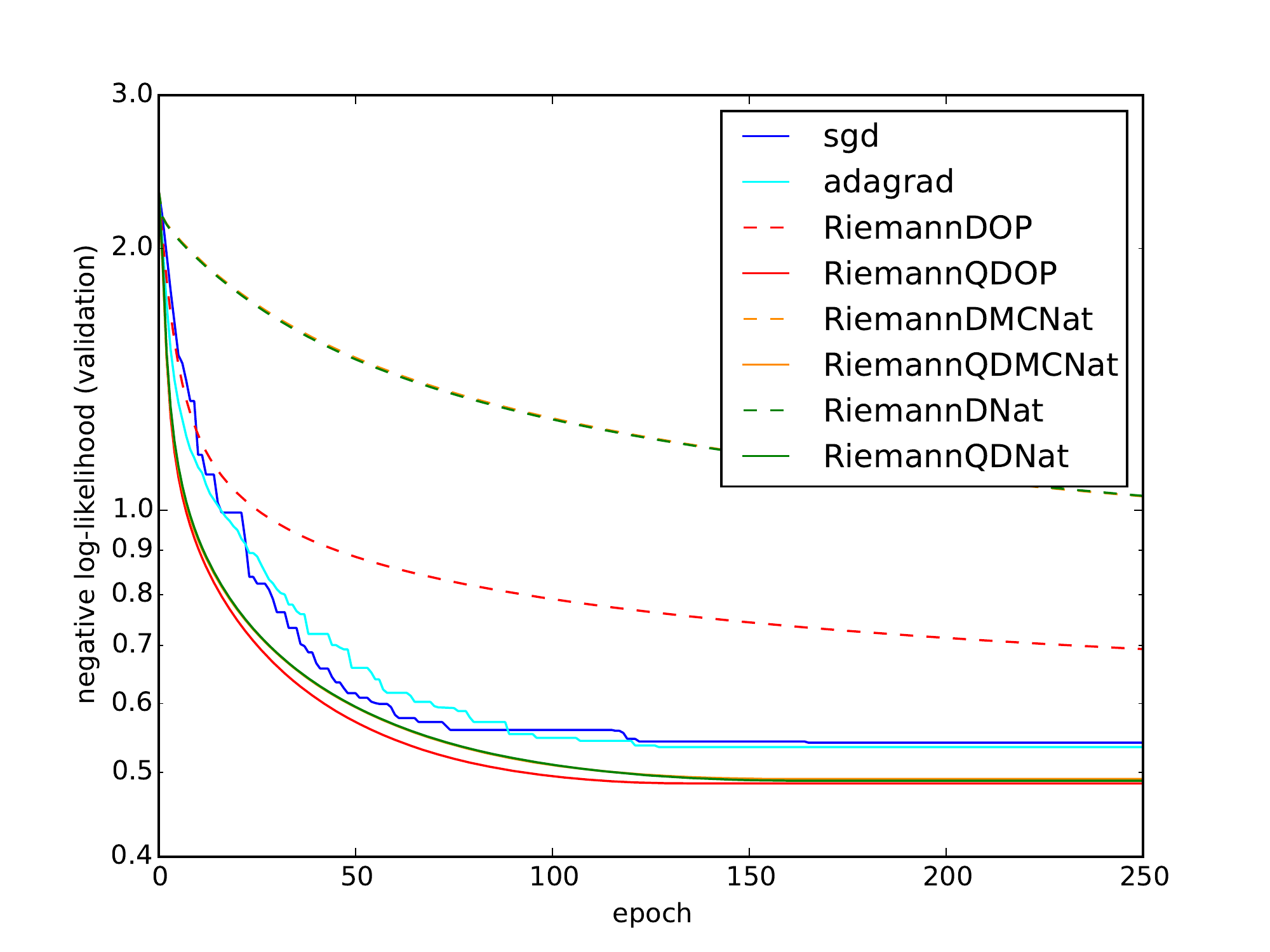}
  \caption{\label{fig:svhn_nll_NN800-800_sigmoid}Classification task on
  SVHN with a non-convolutional 1024-800-800-10 architecture and sigmoid (left) or ReLU (right).
  The network is not regularized.}
\end{figure}

On a more difficult dataset, the permutation invariant SVHN with
grayscale images, 
we observe the same pattern as for the MNIST dataset, with the
quasi-diagonal algorithms leading on SGD and AdaGrad
(\Cref{fig:svhn_nll_NN800-800_sigmoid}).
However, ReLU is severely impacting the \emph{diagonal} approximations of
the Riemannian algorithms:
they diverge for step-sizes larger than $10^{-5}$. This emphasizes once
more the
importance of the quasi-diagonal terms.

\begin{figure}[h]
  \centering
  \includegraphics[scale=0.35]{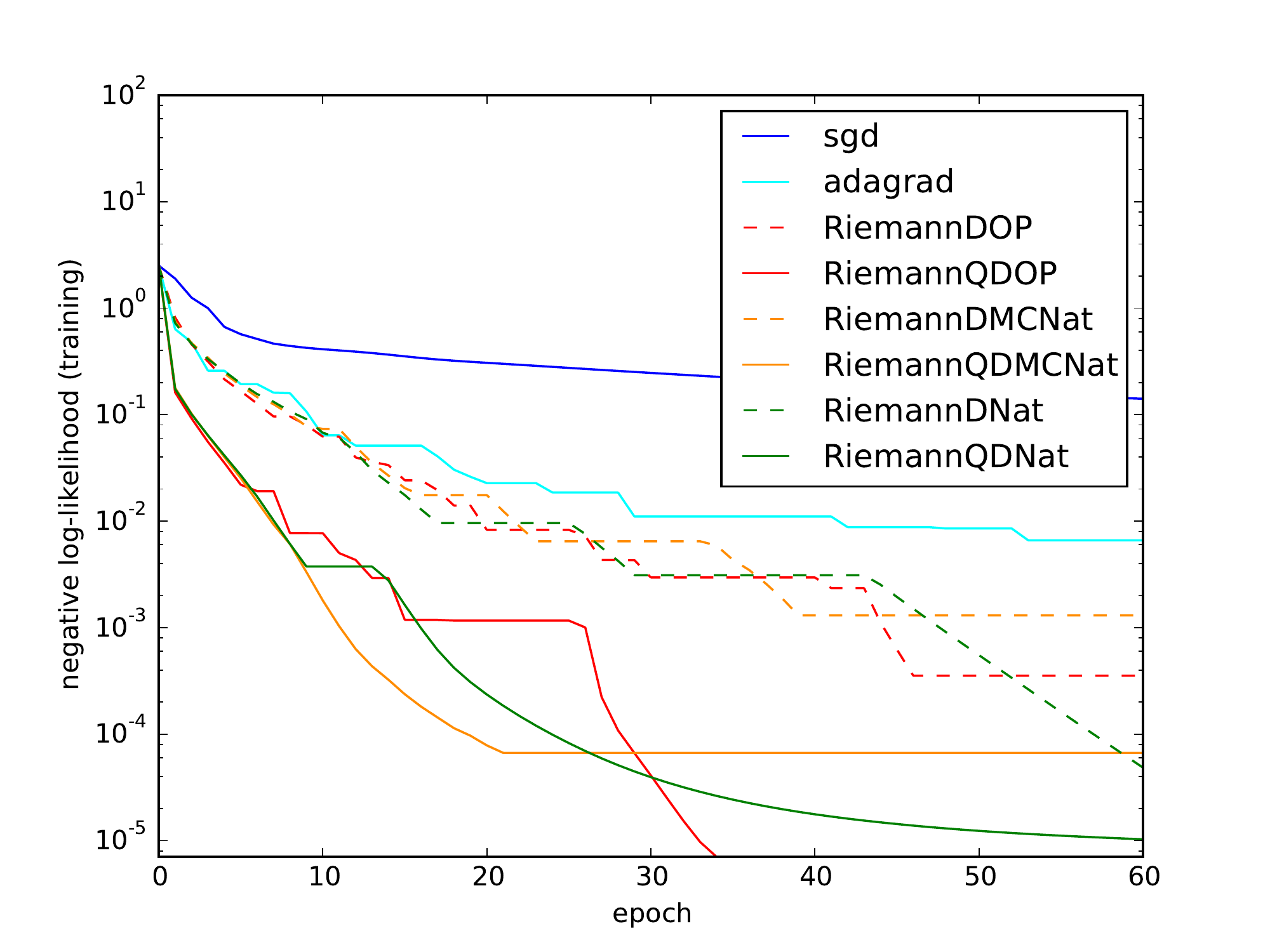}
  \includegraphics[scale=0.35]{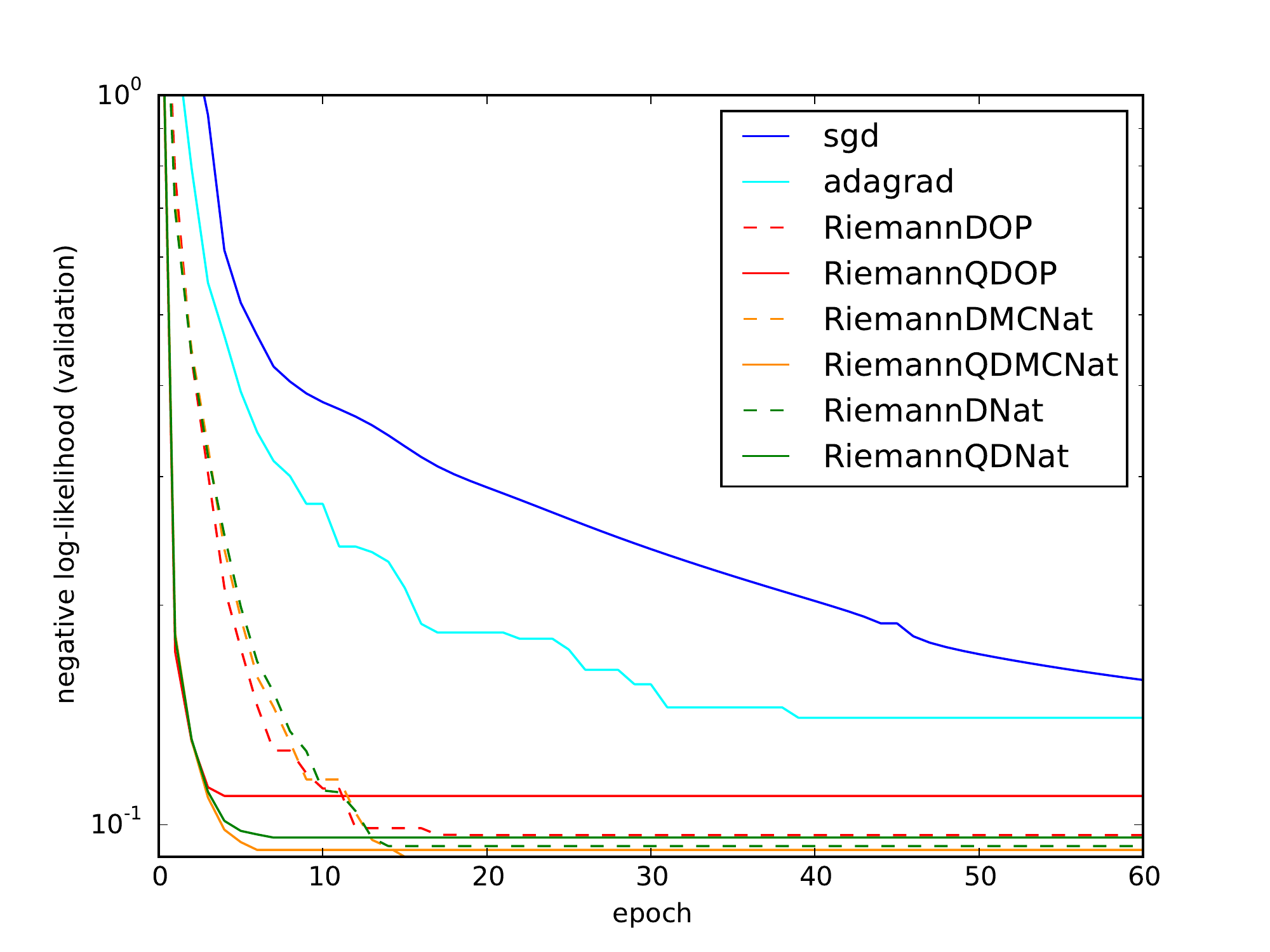}
  \caption{\label{fig:mnist_sp_nll_NN800-800_sigmoid}Classification task
  on MNIST with a 784-2560-1280-640-320-160-80-40-20-10 architecture and sigmoid activation function. We use a sparse network (sparsity=10) with 56310 parameters without regularization. The training trajectory is on the left and the validation trajectory is on the right.}
\end{figure}

We also test the algorithm on a deeper architecture on the MNIST
dataset, with the goal of testing whether Riemannian algorithms handle
``vanishing gradients'' \cite{hochreiter:1991} better.
We use a sparse network with a connectivity factor of 10 incoming weights
per unit.  The connection graph is built from the output to the input by
randomly choosing, for each unit, ten units from the previous layer.
The last output layer is fully connected.  We choose a network
with 8 hidden layers with the following architecture:
2560-1280-640-320-160-80-40-20.  The number of parameters is relatively
small (56310 parameters), and since the architecture is deep, this
should be a difficult problem already as a purely optimization task
(i.e., already on the training set).
From
\Cref{fig:mnist_sp_nll_NN800-800_sigmoid}, we observe three groups: SGD is very
slow to converge, while the
quasi-diagonal algorithms reach quite small loss values on the training
set (around $10^{-4}$ or $10^{-5}$) and perform reasonably well on the validation set. AdaGrad and the diagonal
approximations stand in between.
Once more, quasi-diagonal
algorithms have a steep descent during the first epochs.
Note the various plateaus of several algorithms when they reach very
small loss values; this may be related to numerical issues for such small
values, especially as we used a fixed step size.
Indeed,
most of the plateaus are due to minor instabilities which cause small
rises of the loss values. This may disappear with step sizes tending to
$0$ on a schedule.

Next, we evaluate the Riemannian gradient descents on a regression task,
namely, reconstruction of the inputs.  We use the MNIST dataset and the
{\em faces in the wild} dataset (FACES) \cite{Huang:2007a}, again in a
permutation-invariant non-convolutional setting.  For the FACES dataset,
we crop a border of 30 pixels around the image and we convert it into a
grayscale image.  The faces are still recognizable with this lower
resolution variant of the dataset.
For this dataset, we use an autoencoder with three hidden layers
(256-64-256) and a
Gaussian output.

We also use a dataset of EEG signal
recordings.\footnote{This dataset is not publicly available due to
privacy issues.} These are raw signals captured with 56 electrodes (56
features) with 12 000 measurements.  The goal is to compress the signals
with very few hidden units on the bottleneck layer, and still be able to
reconstruct the signal well.  Notice that these signals are very noisy.
For this dataset we use an autoencoder with seven hidden layers
(32-16-8-4-8-16-32) and a Gaussian output.

In such a setting, the outer product and Monte
Carlo natural gradient are well-suited
(Algorithms~\ref{alg:qdop}--\ref{alg:qdmcnat}), but the exact natural
gradient (Algorithm~\ref{alg:qdnat}) scales like the dimension of the
output and thus is not reasonable (except for the EEG dataset).

\begin{figure}[h]
  \centering
  \includegraphics[scale=0.35]{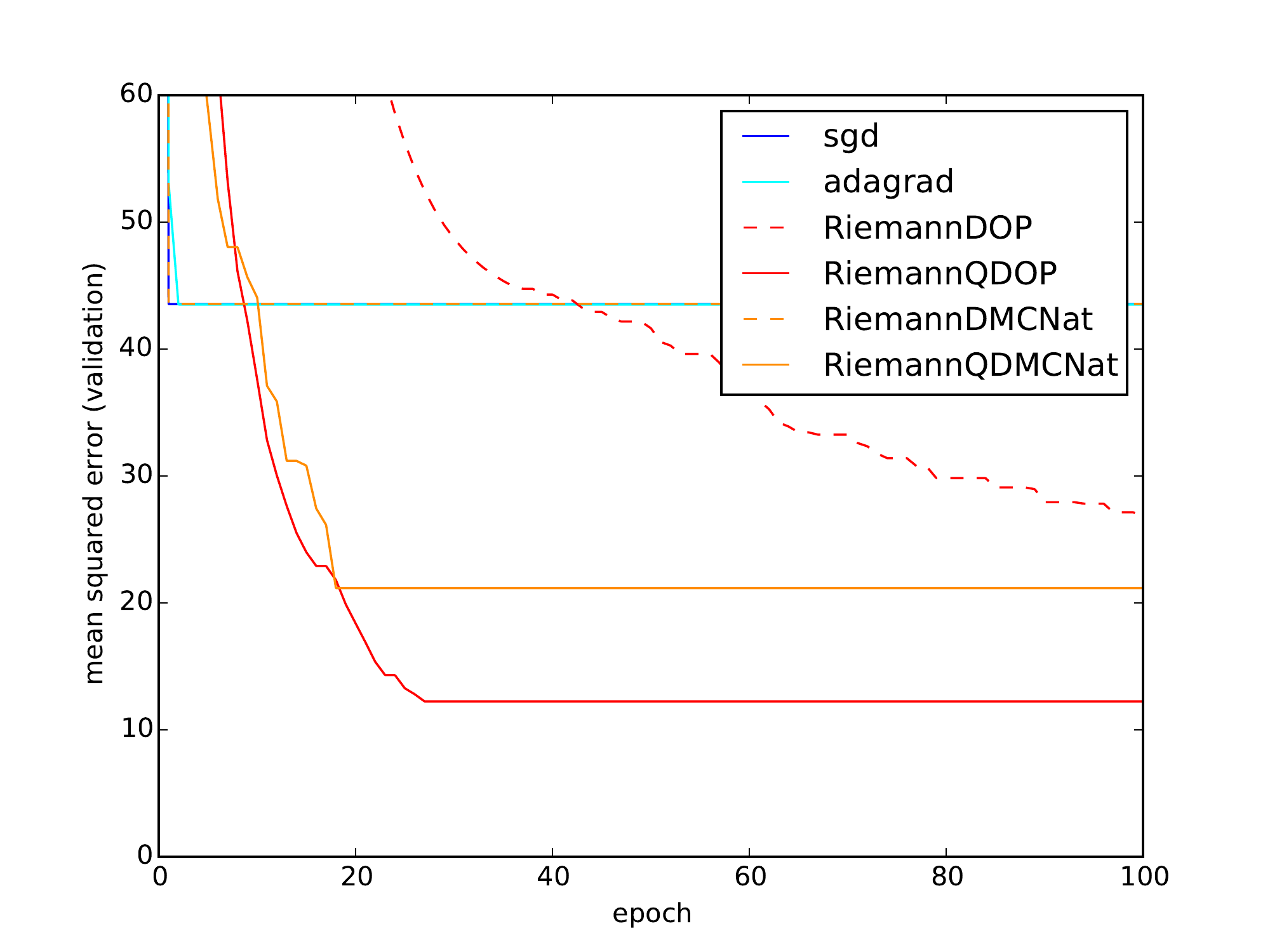}
  \caption{\label{fig:faces_EEG_nll_sigmoid}Reconstruction task on FACES
  (left) with a 2025-256-64-256-2025 architecture.
  The network is not regularized, has
  a sigmoid activation function and a Gaussian output with unit variances.}
\end{figure}

The first experiment, depicted on
\Cref{fig:faces_EEG_nll_sigmoid}, consists in minimizing the mean square
error of the reconstruction for the FACES dataset.  (The mean square error
can be interpreted in terms of a negative log-likelihood loss by defining
the outputs as the mean vector of a multivariate Gaussian with variance
equals to one for every output.)

\begin{figure}[h]
  \centering
  \includegraphics[scale=0.35]{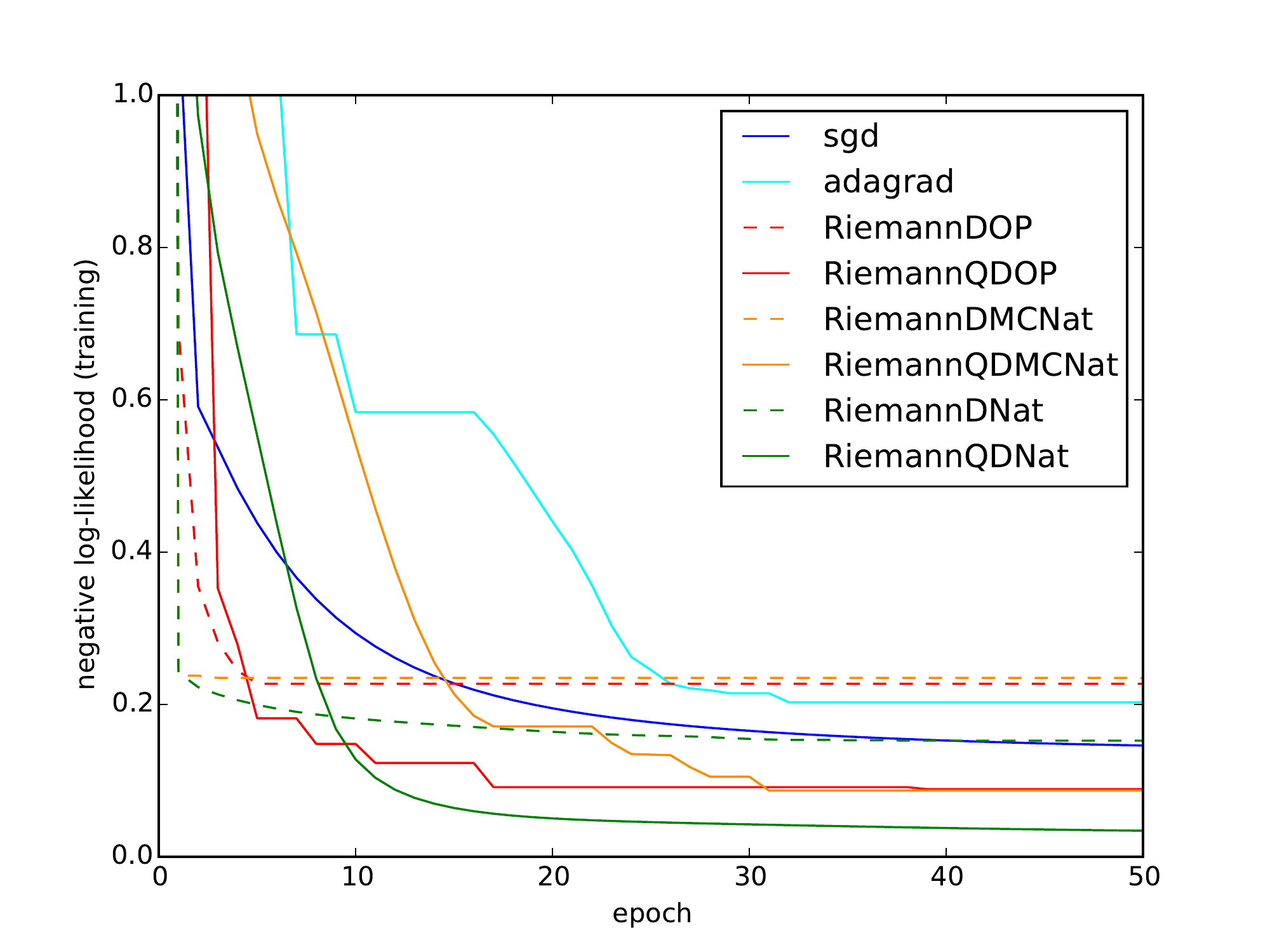}
  \includegraphics[scale=0.35]{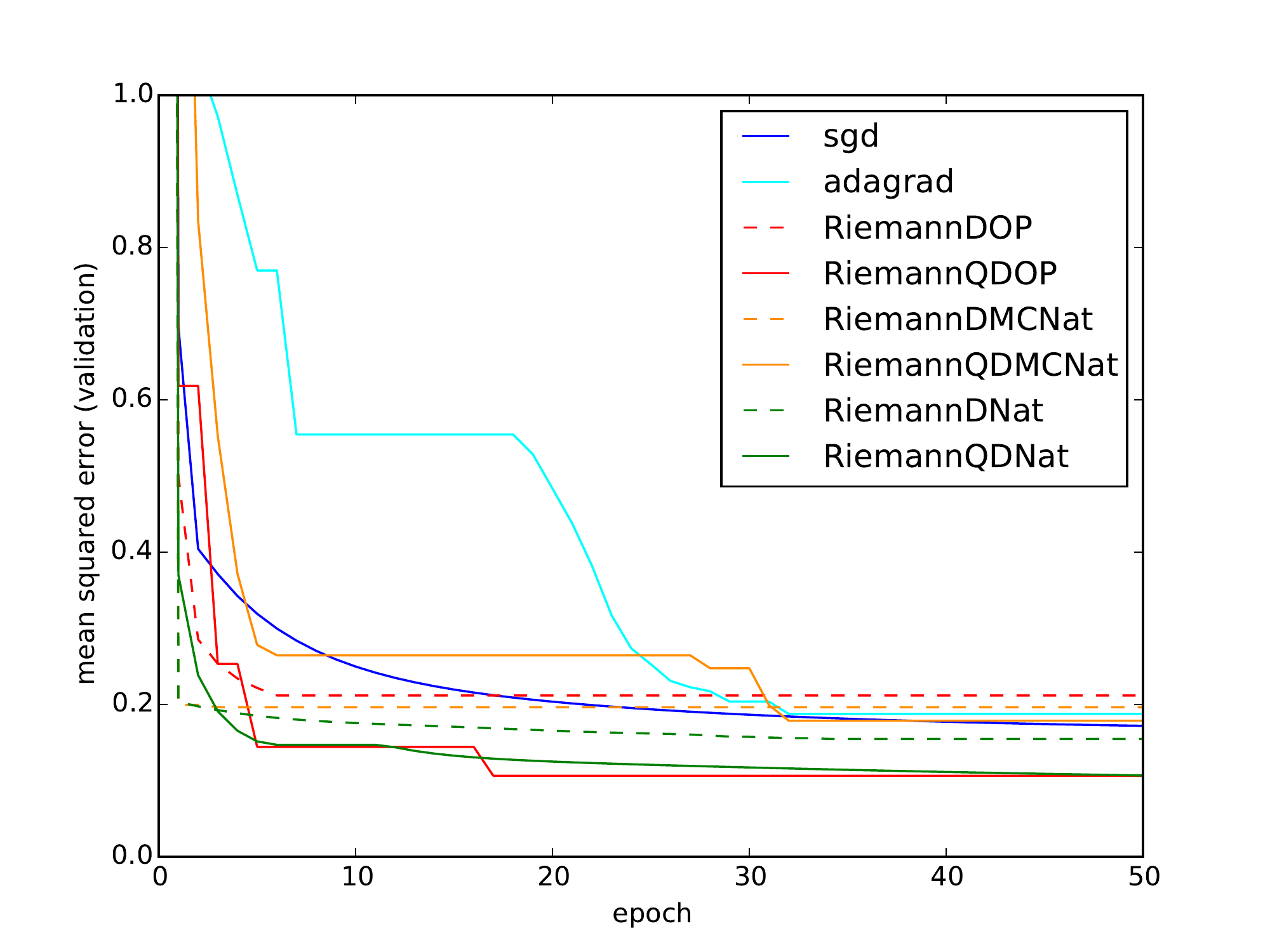}
  \caption{\label{fig:EEG_nll_sigmoid}Reconstruction task on EEG with a
  56-32-16-8-4-8-16-32-56 architecture. The network is not regularized,
  has a sigmoid activation function and a Gaussian output with unit variances.}
\end{figure}

For this experiment, SGD, AdaGrad and RiemannDMCNat are stuck after one
epoch around a mean square error of 45.  In fact, they continue to
minimize the loss function but very slowly, such that it cannot be
observed on the figure.  This behavior is consistent for every step-size
for these algorithms, and may be related to finding a bad local optimum.
On the other hand, with a small step-size, the quasi-diagonal algorithms
successfully decrease the loss well below 45, and do so in few epochs,
yielding a sizeable gain in performance.

On the EEG dataset (\Cref{fig:EEG_nll_sigmoid}), the gain in performance of the quasi-diagonal
algorithms is also notable, though not quite as spectacular as on
FACES.

\begin{figure}[h]
  \centering
  \includegraphics[scale=0.35]{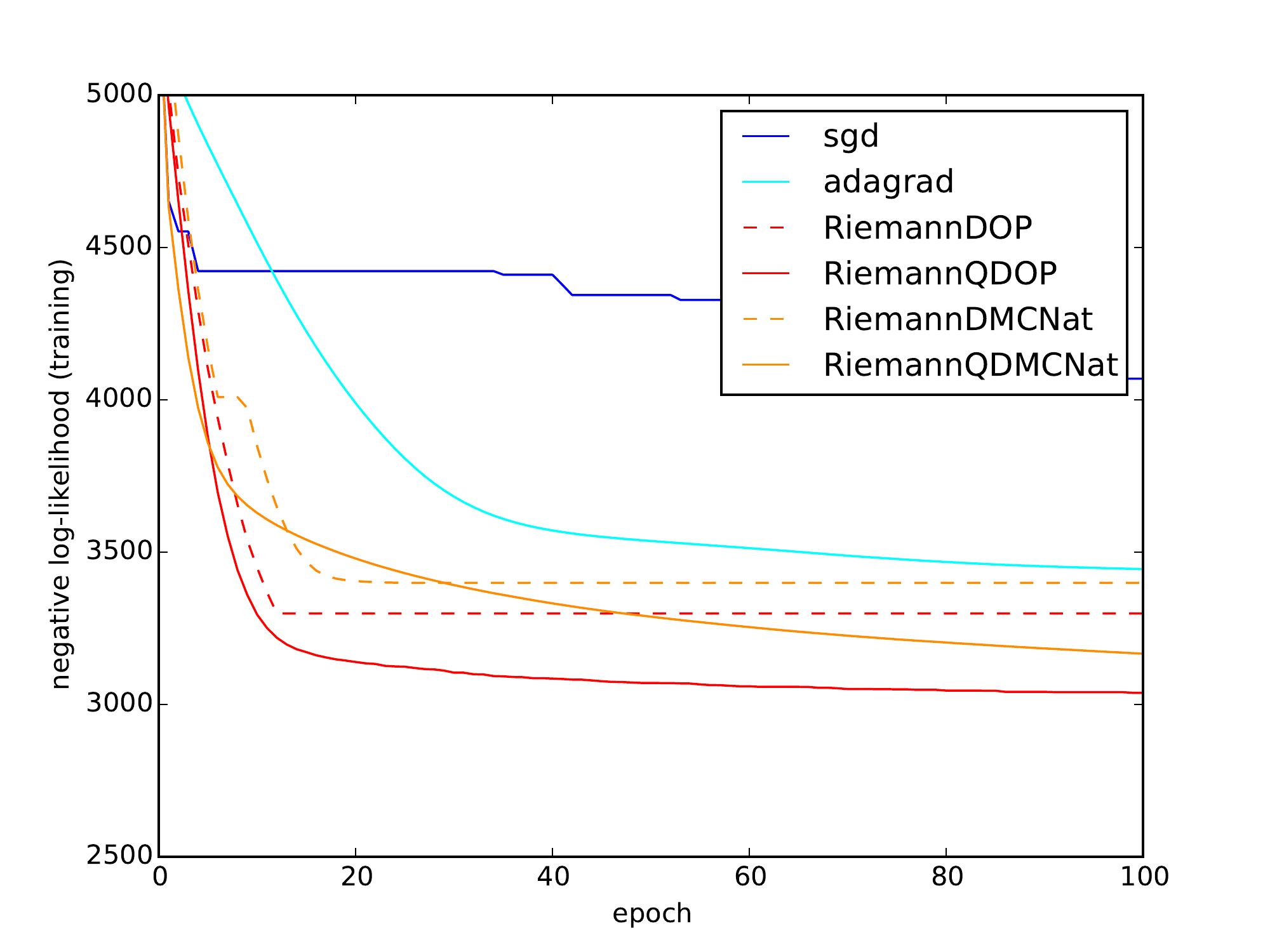}
  \includegraphics[scale=0.35]{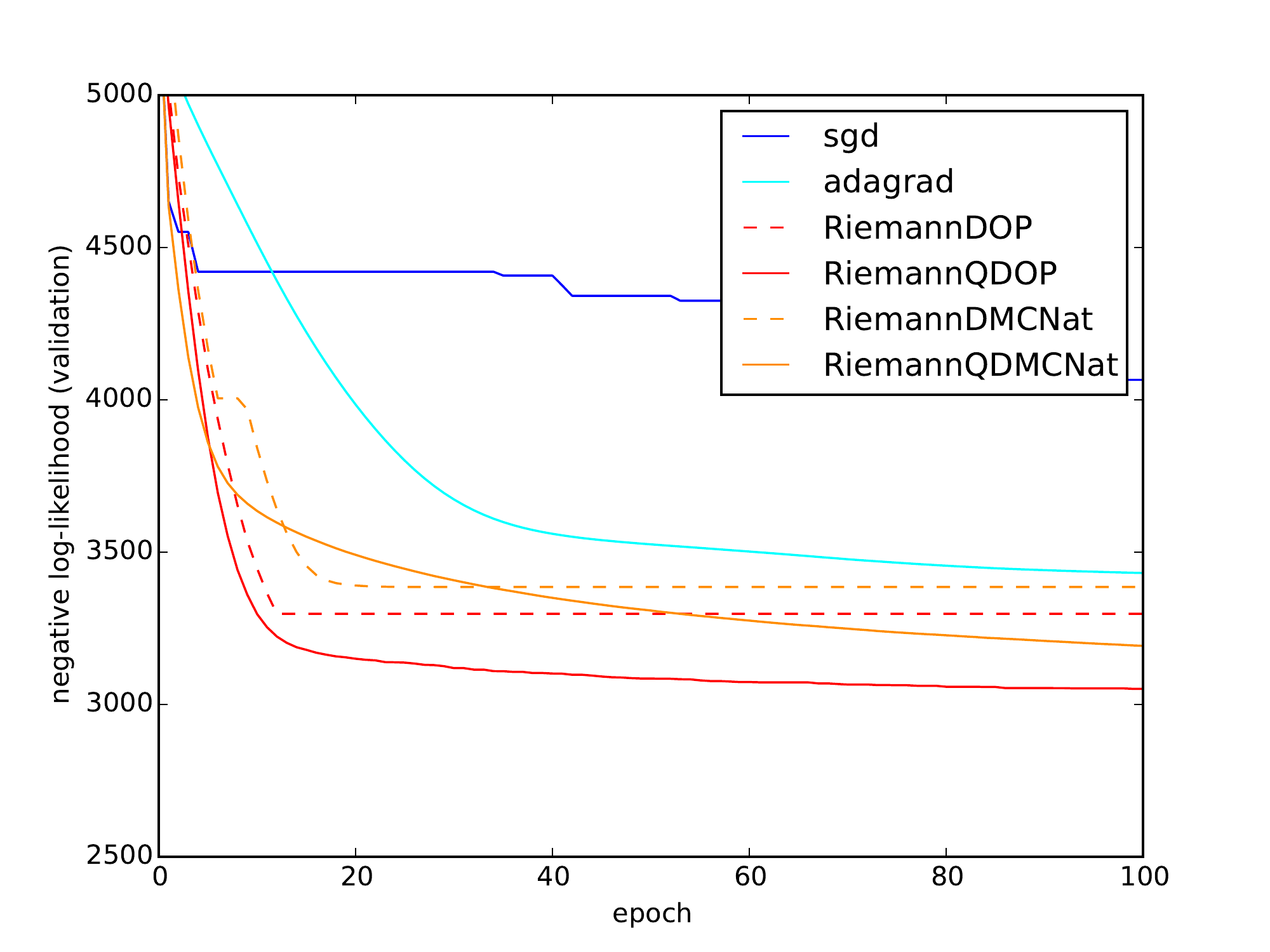}
  \caption{\label{fig:mnist_nll_AE_NN256-64-256_sigmoid}Reconstruction
  task on MNIST with a 784-256-64-256-784 architecture (autoencoder), a
  sigmoid activation function and a Gaussian ouput with learned
  variances.}
\end{figure}

Finally we trained an autoencoder with a
multivariate Gaussian output on MNIST, where the variances of the outputs
are also learned at the same time as the network
parameters.\footnote{As the MNIST data is quantized with $256$ values, we
constrained these standard deviations to be larger than $1/256$.
Otherwise, reported negative log-likelihoods can reach arbitrarily
large negative values when the error becomes smaller than the quantization
threshold.}  As
depicted on \Cref{fig:mnist_nll_AE_NN256-64-256_sigmoid}, the
performances are consistent with the previous experiments.
Interestingly, QDOP is the most efficient algorithm even though the
real noise model over the outputs departs from the diagonal Gaussian
model on the output, so that the OP approximation to the natural gradient
is not necessarily accurate.

\paragraph{Learning rates and regularization.} Riemannian
algorithms are still sensitive to the choice of the gradient step-size
(like classical methods), and also to the numerical regularization term
(the $\epsilon$ in procedure \qdsolve), which was set to
$\epsilon=10^{-8}$ in all our experiments.
The numerical regularization term is necessary to ensure that the metric
is invertible.  However, this term also breaks some invariance and thus,
it should be chosen as small as possible, in the limit of numerical
stability.  In practice, RiemannQDMCNat seems to be more sensitive to
numerical stability than RiemannQDOP and RiemannQDNat.

Moreover, both the Riemannian algorithms and AdaGrad use an additional
hyperparameter $\gamma$, the decay rate used in the moving average of the
matrix $M$ and of the square gradients of AdaGrad. The experiments above
use $\gamma=0.01$. 


\paragraph{Conclusions.}
\begin{itemize}
\item The gradient descents based on quasi-diagonal Riemannian metrics,
including the quasi-diagonal natural gradient,
can easily be implemented on top on an existing framework which computes
gradients, using the routines $\qdrankoneupdate$ and $\qdsolve$. The
overhead with respect to simple SGD is a factor about $2$ in typical
situations.

\item The resulting quasi-diagonal learning algorithms perform quite
consistently across the board whereas performance of simple SGD or
AdaGrad is more sensitive to design choices such as using ReLU or sigmoid
activations.

\item The quasi-diagonal learning algorithms exhibit fast improvement 
in the first few epochs of training, thus reaching their final
performance quite fast. The eventual gain over a well-tuned SGD
or AdaGrad trained for many more epochs can be small or large depending
on the task.

\item The quasi-diagonal Riemannian metrics widely outperform their
diagonal approximations (which break affine invariance properties). The
latter do not necessarily perform better than classical algorithms. This
supports the specific influence of invariance properties for performance,
and the interest of designing algorithms with such invariance properties
in mind.

\end{itemize}



\bibliographystyle{alpha}
\bibliography{library}
\end{document}

%% file: header.tex
\newcommand{\transp}[1]{#1^{\!\top}\!}

\def\R{{\mathbb{R}}}

\renewcommand{\geq}{\geqslant}

\newcommand{\deq}{\mathrel{\mathop:}=}

\newcommand{\from}{\colon} 

\def\eps{\varepsilon}
\renewcommand{\epsilon}{\varepsilon}
\renewcommand{\phi}{\varphi}

\let\oldPr\Pr
\renewcommand{\Pr}{\oldPr\nolimits}
\newcommand{\E}{\mathbb{E}}

\DeclareMathOperator{\Id}{Id}

\newcommand{\norm}[1]{\left\|#1\right\|}

\DeclareMathOperator*{\argmin}{arg\,min}